\documentclass[10pt,journal,compsoc]{IEEEtran}
\usepackage{tikz}
\usepackage{tabularx}
\usepackage{array}
\usepackage{booktabs}

\newcolumntype{Y}{>{\raggedleft\arraybackslash}X}
\newcolumntype{C}[1]{>{\centering\arraybackslash}p{#1}}

\usepackage{xurl}
\usepackage[hidelinks]{hyperref}
\ifCLASSOPTIONcompsoc
  \usepackage[nocompress]{cite}
\else
  \usepackage{cite}
\fi
\usepackage{amsmath}
\usepackage{amssymb}
\newcommand{\cmark}{\checkmark}
\newcommand{\xmark}{--}

\usepackage{algorithm}
\usepackage{algorithmic}
\usepackage{array}
\ifCLASSOPTIONcompsoc
 \usepackage[caption=false,font=footnotesize,labelfont=sf,textfont=sf]{subfig}
\else
 \usepackage[caption=false,font=footnotesize]{subfig}
\fi
\usepackage{fixltx2e}
\usepackage{stfloats}
\begin{document}

%don't want date printed
\date{}
% \title{Community Hiding from Graph Neural Networks}
% \title{Hiding Communities from Graph Neural Networks}
\title{Community Concealment from Graph Neural Networks}
\author{\IEEEauthorblockN{Dalyapraz Manatova\IEEEauthorrefmark{1},
Pablo Moriano\IEEEauthorrefmark{2} and 
L. Jean Camp\IEEEauthorrefmark{1}}\\
\IEEEauthorblockA{\IEEEauthorrefmark{1}  UNC Charlotte, Charlotte, NC, USA}
\IEEEauthorblockA{\IEEEauthorrefmark{2} Oak Ridge National Laboratory, Oak Ridge, TN, USA}
}

% \markboth{Journal of \LaTeX\ Class Files,~Vol.~14, No.~8, August~2015}%
% {Shell \MakeLowercase{\textit{et al.}}: Bare Demo of IEEEtran.cls for Computer Society Journals}


%  Authors in ACM format
% \author{
% {\rm Dalyapraz Manatova}\\
% Indiana University
% \and
% {\rm Pablo Moriano}\\
% Oak Ridge National Laboratory
% \and
% {\rm L. Jean Camp}\\
% UNC Charlotte
\IEEEtitleabstractindextext{%
\begin{abstract}
Graph neural networks (GNNs) enable powerful unsupervised learning of communities. However, such inference may inadvertently expose sensitive group structures, critical clustered patterns, or collective behaviors, raising concerns about sensitive group-level privacy. In social and critical infrastructure networks, unauthorized community inference can reveal coordinated asset groups, operational hierarchies, and system dependencies that may be exploited for reconnaissance or profiling. We study a defensive setting in which a network (or defender) operator seeks to conceal a community of interest while making only small, utility-preserving modifications to the network. Our analysis shows that community concealment depends on two measurable factors: the connectivity at the community boundary and the feature similarity between the protected community and its neighbors. Guided by these observations, we introduce Feature-Community-guided DICE (FCom-DICE), a perturbation strategy built on DICE (Disconnect Internally Connect Externally) that rewires a set of structurally influential edges and adjusts node features to reduce the distinctiveness exploited by GNN message passing. 
Across synthetic benchmarks and real network graphs such as Facebook, Wikipedia, and Bitcoin Transactions, FCom-DICE consistently outperforms structure-only DICE under the same perturbation budgets. The largest improvements are observed for communities that are weakly connected to the rest of the network and well separated in feature space. These gains are achieved while preserving key structural and feature characteristics of the original network.
% Overall, it achieves median relative concealment improvements of approximately 20–45\% on synthetic and real networks. 
These results demonstrate the effectiveness of feature-aware perturbations for reducing the recoverability of targeted communities under GNN-based community inference.
% Our source code is publicly available at: https://github.com/dalyapraz/gnn-clustering-attack-defense/tree/artifact
\end{abstract}

% Note that keywords are not normally used for peer review papers.
}

\maketitle

\IEEEdisplaynontitleabstractindextext

% \fi
\IEEEpeerreviewmaketitle

\section{Introduction}

\begin{figure}[!ht]
    \centering
    \includegraphics[width=\linewidth]{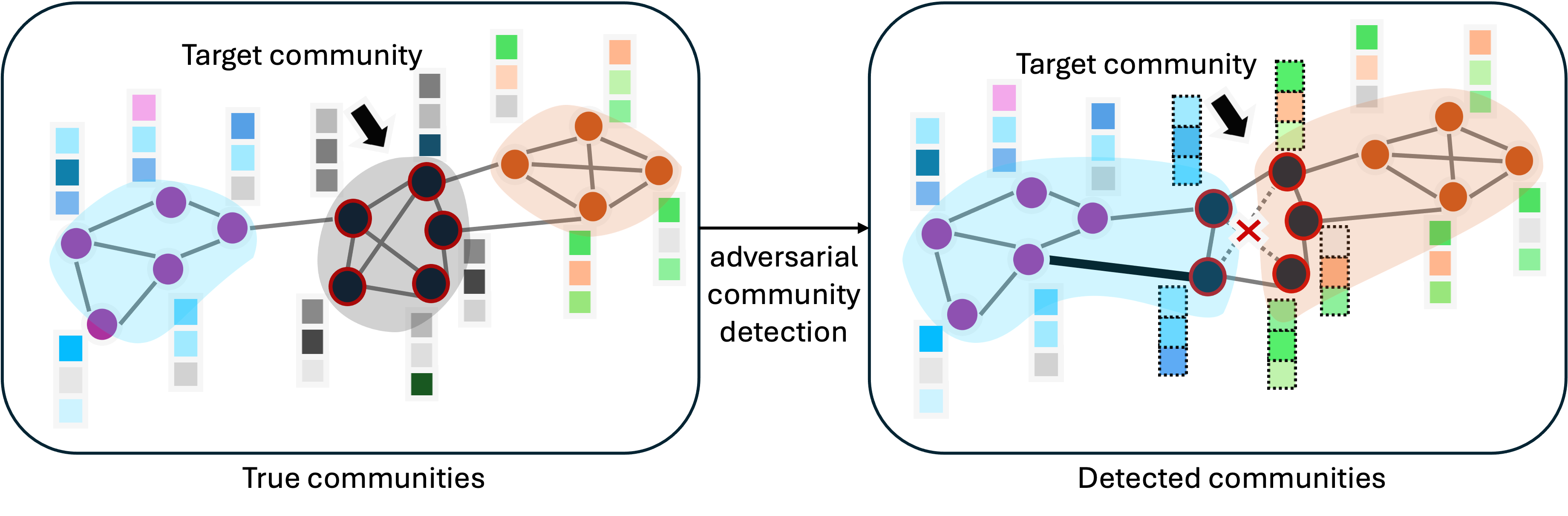}
    \vspace{-1em}
    \setlength{\abovecaptionskip}{0pt}
    \caption{Adversarial community detection scenario. Left: the original network containing several communities, including one target community that must remain concealed. Right: the output of a GNN used by an adversary to infer community structure. The defender’s goal is to slightly modify the target community so that the adversary's GNN fails to recover it correctly.}
    \vspace{-1em}
    \label{fig:concept_com_hiding}
\end{figure}

Community detection is a fundamental task in network science, revealing groups of densely connected nodes that often share similar functions~\cite{fortunato_community_2010, girvan_community_2002, radicchi_defining_2004}, such as coordination and collective behaviors in social networks~\cite{pacheco_uncovering_2021, manatova_understand_2024}, or interdependence of assets in critical infrastructure~\cite{moriano_community-based_2019}. 
% In social networks, it helps uncover patterns of coordination and collective behavior~\cite{pacheco_uncovering_2021}, whereas in critical infrastructure systems, it reveals clusters of assets whose interdependence affects operational resilience~\cite{moriano_community-based_2019}. 
Recent graph neural networks (GNNs) have advanced community detection by integrating structural and feature information~\cite{park_unsupervised_2020, tsitsulin_graph_2023, goel_community_2026}. 
% Yet this same capability introduces new security and privacy concerns. 
% When exploited by adversaries, GNN-based community detection can expose sensitive relationships, operational hierarchies, or control dependencies that were previously difficult to infer.
However, this capability introduces a group-level privacy risk, as GNN-based community detection can infer sensitive collective relationships, operational hierarchies, and system dependencies even when no individual node or edge reveals the group in isolation.
% This exposure constitutes a form of group-level privacy leakage because 
% With GNN-based community detection, sensitive collective structure can be inferred even when no individual node or edge reveals the group in isolation.
%% Adversarial Threat and Defender Perspective
% Adversaries can leverage GNN-based community detection to extract insights from both network topology and node attributes, uncovering clusters that provide extensive information regarding the networks. 
% In social networks, such analysis can uncover influence groups or coordinated information campaigns; in cyber-physical or enterprise systems, it may expose operational structures that were intended to remain private. 
% In social and critical infrastructure networks, unauthorized community inference can reveal 
Adversaries can leverage GNN-based community detection to extract insights about coordinated asset groups, clustered behavioral patterns, or system dependencies that are valuable for profiling or reconnaissance, from both network topology and node attributes.
Figure~\ref{fig:concept_com_hiding} illustrates this setting: 
% a defender operates a network or a set of nodes (i.e., community) in a network that contains one or more communities of interest, while an adversary applies a GNN to infer them. To counter this privacy risk, the
a defender seeks to modify the graph just enough to ensure that the GNN run by the adversary fails to correctly recover the targeted community~\cite{chang_community_2024, tekin_qualitative_2024}. 
% This defensive objective differs from generic privacy attacks or random perturbations because the defender must balance concealment with the preservation of structural and semantic fidelity.

% \paragraph{Problem Formulation.}
%Target group of nodes (target community) in the network with node features: true communities (left) and communities detected by an adversary---target community is split and not recovered (right). 

\noindent\textbf{Problem Formulation.}
Community concealment under GNN-based detection fundamentally differs from classical community concealment~\cite{fionda_community_2016} because, rather than inferring communities based on structure only, GNNs infer communities from both topology and node features through message passing~\cite{goel_community_2026, tsitsulin_graph_2023}. 
Consequently, the defender aims to reduce the ability of the GNN to correctly identify a targeted community by introducing small, controlled perturbations to the network, while preserving key graph properties, such as overall network community structure,
degree distribution, and node-feature semantics.
% since the modified graph must remain functional for legitimate analytics and operations. 
Although DICE~\cite{waniek_hiding_2018} provides an effective structure-only baseline for concealing communities by removing internal links and adding external ones, it does not account for feature information that is used in GNN-based methods.
% its design assumes purely structural detection. 
% GNN-based methods exploit both structure and attributes through message passing, making the community concealment substantially more complex~\cite{goel_community_2026, tsitsulin_graph_2023}. 
This motivates two fundamental questions: \textit{Can a group of nodes (a community) be intentionally concealed from graph-learning-based unsupervised clustering?} 
And if so, \emph{what characteristics govern the concealment of a community under a GNN-based community detection algorithm?} Naive extensions of DICE often fail to mislead such models, since feature homophily and multi-hop aggregation in GNNs can still reveal latent community boundaries. 
These observations define the central challenge: \textit{how to reduce the recoverability of a targeted community under GNN-based community inference while operating within a limited perturbation budget.}

%% Key Insight and Approach Intuition
% Our analysis begins with the observation that the success of community concealment depends on measurable structural and feature-based properties of the network. 
Through controlled experiments on graphs, we find that the ability to conceal a community from a GNN is strongly influenced by two factors: the ratio of external to internal connections and the similarity of node features between the targeted community and its neighboring ones. Communities that are weakly connected to the rest of the network or whose features differ sharply from their surroundings tend to be easily rediscovered after perturbation, while those with moderate boundary connectivity and overlapping features are more likely to remain hidden. These findings reveal that community concealment is not uniform but governed by predictable structural and feature-space characteristics. Guided by this insight, we develop \textit{FCom-DICE (Feature-Community-guided DICE)}, a concealment strategy that extends DICE~\cite{waniek_hiding_2018} with feature-aware perturbations, rewiring a small number of edges and minimally adjusting node features to disrupt GNN message-passing~\cite{goel_community_2026}. Figure~\ref{fig:concept_com_hiding} (right) conceptually illustrates this intuition: modifying graph structure at community boundaries while reducing feature-space separability can alter the representation learned by a GNN, causing the targeted community to be incorrectly recovered or merged with others.

%%Evaluation Scope and Findings 
\noindent\textbf{Contributions.} 
We evaluate the proposed concealment strategy FCom-DICE on synthetic and real networks, including Facebook, Wikipedia, and Bitcoin transactions.  Synthetic experiments control variation of community separability, feature similarity, and perturbation budgets, while the real graphs provide realistic structural and attribute distributions. 
Across all settings, incorporating node features into the perturbation strategy consistently improves or matches the effectiveness of the structure-only DICE baseline~\cite{waniek_hiding_2018} under the same perturbation constraints. 
Although the magnitude of improvement varies with the GNN architecture, the results consistently demonstrate that community concealment in attributed networks is governed by both structural and feature-space separability.
% The median relative improvement ranges from approximately 20\% to 45\% under the best-performing detector across synthetic and real datasets, demonstrating that feature-guided perturbations consistently improve the concealment of targeted communities under GNN-based community inference. 
These gains are achieved while preserving the overall network community structure and degree distribution. 
Moreover, the perturbations remain statistically difficult to distinguish from the original graph: the modified networks satisfy established degree-distribution unnoticeability criteria for perturbation budgets up to 15\%, while feature perturbations remain statistically indistinguishable with modifications of up to 20 nodes (20\% of the size of the largest community) under the largest feature-space separation considered.
% key utility properties: the modified graphs maintain comparable degree distributions, community sizes, and node-feature semantics. 
Together, these results demonstrate that incorporating feature information into the perturbation process provides a consistent advantage over structure-only concealment under GNN-based community detection~\cite{goel_community_2026, tsitsulin_graph_2023}.

%%Contributions Paragraph
In summary, this work makes the following contributions. First, we formulate the problem of defensive community concealment under GNN-based community detection, where a defender seeks to conceal a targeted community through small, utility-preserving perturbations. Second, through controlled experiments, we identify two measurable factors that govern community concealment: the ratio of external to internal connectivity and the similarity of node features between neighboring communities. Third, we introduce FCom-DICE, a defense strategy that extends DICE by incorporating feature-aware edge rewiring and budgeted feature adjustment to disrupt GNN message passing. Fourth, we conduct a comprehensive evaluation on synthetic networks and real ones, demonstrating that FCom-DICE consistently outperforms the structure-only DICE baseline, with the largest gains observed when communities are loosely connected to the rest of the network and well separated in feature space.
Finally, we discuss the implications of these findings for protecting high-value communities in operational and critical infrastructure networks.
To support reproducibility and further research, we make our implementation, experimental pipeline, and evaluation code publicly available at \url{https://github.com/dalyapraz/gnn-community-concealment}
\section{Related Work}
\label{sec:related_work}

\begin{table*}[!htb]
\caption{Comparison of Related Work on Community Concealment}
\label{tab:community_concealment_work}
\centering
% \scriptsize
\setlength{\tabcolsep}{4pt}
\renewcommand{\arraystretch}{1.18}
\begin{tabularx}{\textwidth}{@{}
p{3.2cm}
C{1cm}
C{1.25cm}
C{1.25cm}
X
@{}}
\toprule
\textbf{Author, Year [Ref.] (Method)} &
\textbf{Against GNN} &
\multicolumn{2}{c}{\textbf{Concealment Perturbs}} &
\textbf{Paper's Main Objective} \\
\cmidrule(lr){3-4}
&
&
% \textbf{Edge Addition} & \textbf{Edge Deletion} & \textbf{Feature Change} & \\
\textbf{Graph Structure} & \textbf{Node Features} & \\
\midrule
Nagaraja, 2010~\cite{nagaraja_impact_2010} &
\xmark & \cmark & \xmark &
Evaluation of privacy-oriented target group concealment from adversarial community detection via edge addition.\\

Waniek et al., 2018~\cite{waniek_hiding_2018} (DICE) &
\xmark & \cmark & \xmark &
Node and community concealment from classical structure-based community detection via edge addition/deletion; introduces $M_1$ and $M_2$ concealment metrics. \\

Fionda and Pirr\`o, 2018~\cite{fionda_community_2018} &
\xmark & \cmark & \xmark &
Community deception from classical structure-based community detection via edge addition/deletion; introduces deception/safeness metrics. \\

Li et al., 2020~\cite{li_adversarial_2020} (CD-Attack) &
\cmark & \cmark &  \xmark &
Community concealment method from graph-learning-based community detection via edge addition/deletion. \\

Chen et al., 2021~\cite{chen_multiscale_2021} (EPA) &
\xmark & \cmark & \xmark &
Multiscale concealment method of target nodes, communities, and global community structure via edge addition/deletion. \\

Mittal et al., 2021~\cite{mittal_hide_2021} (NEURAL) &
\xmark & \cmark & \xmark &
Community concealment method from classical structure-based community detection with limited perturbations via edge addition/deletion. \\

Liu et al., 2022~\cite{liu_community_2022} (GCH) &
\xmark & \cmark &  \xmark &
Community concealment method using graph-autoencoder-guided structural perturbation via edge addition/deletion. \\

Zhang et al., 2024~\cite{zhang_community_2024} (ComDeceptor) & \xmark & \cmark & \xmark & 
Community concealment method using Laplacian-spectrum-guided structural perturbation via edge addition/deletion. \\

Chang et al., 2024~\cite{chang_community_2024} &
\xmark & \cmark &  \xmark &
Community concealment method from classical structure-based community detection via edge addition/deletion. \\

Pirr\`o, 2024~\cite{pirro_community_2024} (nDec) &
\xmark & \cmark & \xmark &
Community concealment method from classical structure-based community detection via node addition/deletion.\\

Fionda and Pirr\`o, 2024~\cite{fionda_community_2024} (AttSafDec) &
\xmark & \cmark & $^{\star}$ &
Attribute-aware community concealment method from classical structure-based community detection. \\
\textbf{This paper} (FCom-DICE) &
\cmark & \cmark &  \cmark & 
Evaluation of factors driving community concealment from GNN-based community detection; factor-informed concealment method via edge and node feature perturbations. \\

\bottomrule
\end{tabularx}
\footnotesize{\textit{Note:} ``Against GNN'' indicates that the concealment setting is designed for or explicitly evaluated against GNN-based community detection.}
\footnotesize{\textit{$^{\star}$}The method uses node attributes to guide concealment but does not directly modify node features.}
\end{table*}

Our research builds upon prior work in (i) community concealment for classical methods and (ii) community concealment for GNN-based clustering. While the GNN literature on adversarial attacks is sizable, it primarily targets node classification~\cite{zugner_adversarial_2018, xi_graph_2021} or link prediction~\cite{wu_linkteller_2022}. Concealment from unsupervised GNN clustering (community detection) remains largely underexplored, with existing approaches relying on structural perturbations. We review closely related work and summarize how these studies compare to our contribution in Table~\ref{tab:community_concealment_work}.

\noindent\textbf{Community Concealment.}
% The robustness of community detection methods has been studied in the literature across three primary categories, distinguished by the scale of the manipulation: target node concealment~\cite{chen_fast_2018}, target community concealment~\cite{nagaraja_impact_2010}, and global community structure attacks~\cite{wei_robustness_2024, chen_ga-based_2019, zhao_self-adaptive_2023}. 
Community concealment has been studied at the level of individual nodes~\cite{chen_fast_2018}, entire communities~\cite{nagaraja_impact_2010}, or global community structure~\cite{wei_robustness_2024, chen_ga-based_2019, zhao_self-adaptive_2023}. 
In this work, we focus on \emph{target community concealment} from being recovered as a coherent cluster.
% i.e., dispersing a single community so it is not recovered as a coherent cluster.
Early work in this space by Nagaraja ~\cite{nagaraja_impact_2010}, showed that adding a small number of edges to the high-centrality nodes can provide privacy to a group of nodes, reducing their recoverability as a cluster. 
% The study showed that strategically placed cross-partition links reduce an eavesdropper’s ability to infer community structure, thereby enhancing privacy for the defended group.
Subsequent work formulated concealment as a budgeted edge perturbation optimized using objectives, such as deception, safeness~\cite{fionda_community_2018, chen_community_2021}, entropy for genetic search~\cite{chen_multiscale_2021}, permanence~\cite{mittal_hide_2021}, and escape or dispersion scores~\cite{chang_community_2024}. 
A widely used baseline is DICE~\cite{waniek_hiding_2018}, a modularity-motivated heuristic that randomly deletes intra-community edges and adds cross-community edges without optimizing a specific score. 

Recent work has extended community concealment to node addition and deletion~\cite{pirro_community_2024} and attribute-aware objectives that use node attribute similarities to guide structural perturbations~\cite{fionda_community_2024}.
While these studies show the importance of node attributes, they
% A recent work by Pirr\'o~\cite{pirro_community_2024} explores a node-centric approach by removing and adding nodes in the community structure to conceal a community. 
% Fionda and Pirr\`o~\cite{fionda_community_2024} further extend the concealment to attributed networks by converting node attribute similarities to edge weights and formulating an attribute-aware concealment objective function. 
% Their findings show that attribute-oblivious concealment methods perform poorly on attributed networks, highlighting the need to account for node attributes when concealing communities. However, these methods 
remain focused on classical community detection and do not directly study feature perturbations against GNN-based community inference. 
% This motivates our focus on GNN-based methods of detection, where node features significantly contribute to the message-passing process in community detection. 

We use DICE as the primary structure-only baseline because its transparent, budget-controlled rewiring isolates the contribution of feature information to community concealment, without introducing optimization objectives.
% Unlike optimization-based methods that introduce additional objectives or search procedures, DICE directly reduces internal connectivity and increases external connectivity without using node attributes. 
This makes it a suitable reference for determining whether feature-guided edge selection and feature modification provide benefits beyond topology-only perturbation

% \noindent\textbf{Adversarial Attacks on GNN.}
% netattack 
% feature attack on classification
% cite some papers from Usenix, S&P and PETS maybe?

\noindent\textbf{Community Concealment from GNN.}
Research work that explicitly targets community concealment from graph learning is sparse.
CD-Attack~\cite{li_adversarial_2020} and GCH~\cite{liu_community_2022} perturb graph topology only to reduce community recoverability using surrogate GNN models, but neither considers feature perturbations. 
% One of the closely related works is an adversarial attack on graph-learning-based community detection known as CD-Attack~\cite{li_adversarial_2020}. The method trains a surrogate GNN community detector with a normalized-cut loss and perturbs edges (addition/deletion) around a target group of nodes via a constrained graph generator. 
% A related approach, GCH~\cite{liu_community_2022}, uses a graph autoencoder to select structural edits that suppress a community’s recoverability. Note that in both methods the perturbation is topology-only.
Recent evaluations of GNN robustness for community detection show that both edge and feature perturbations degrade clustering quality across different GNN architectures, but these works are diagnostic rather than prescriptive methods for community deception or concealment~\cite{goel_community_2026}. 
This leaves a gap for methods that utilize structure and features to conceal communities from GNN-based community detections. We address this gap by 
% Therefore, the goal of this work is to fill this gap by 
investigating what factors drive concealment of a target community from GNN-based community detection and devising a heuristic strategy for target community concealment.

\section{Background}
\label{sec:background}
We begin with preliminary definitions and background to provide context for the problem investigated in this paper.

\subsection{Graphs}
% All the graphs and feature definitions go in here.
A graph is defined as $G=(V,E)$, where $V = (v_1,...,v_n),$ $|V|= n$ is the set of nodes and $E\subseteq V \times V$, $|E| = m$ is the set of edges. Each node $v \in V$ can also have an associated feature vector $\mathbf{x_v} \in \mathbb{R}^d$. We denote the collection of all node features by $ \mathbf{X} = \{\mathbf{x_1}, \mathbf{x_2}, \ldots, \mathbf{x_n}\} $, where $\mathbf{x_i} \in \mathbb{R}^d $ is a $d $-dimensional feature vector corresponding to node $v_i$.
A representation of edges between nodes in a matrix form is defined as an $n \times n$ adjacency matrix of $G$, denoted by $\mathbf{A}$, where $\mathbf{A}_{ij}$ = 1 if and only if $\{v_i, v_j\} \in E$ and otherwise entries of $\mathbf{A}$ are equal to 0.

\subsection{Community Detection}
% Brief description of community detections a measure of community quality
A \textit{community} (or cluster), in the graph context, refers to a subgraph $C_i$ of a graph $G$ where nodes are densely connected inside $C_i$ and loosely connected with nodes from $C_j$, where $i\neq j$ and~$C_i \cap C_j = \emptyset$~for~all~$i, j$~\cite{fortunato_community_2010}. Here, we assume non-overlapping communities. A community detection algorithm seeks to partition the graph $G$ into such clusters (i.e., subgraphs) $\mathcal{C} = \{C_i,...,C_k\}$.
The quality of such partitions is often measured by \textit{modularity}, which quantifies the deviation of the intra-cluster edges in the graph $G$ from the expected number of edges in the random graph~\cite{newman_modularity_2006}.  
In a random graph, nodes $v_i$ and $v_j$ with degrees $\hat{k}_i$ and $\hat{k}_j$ are connected with probability $\hat{k}_i \hat{k}_j/2m$. Then modularity is defined as:
\begin{equation}
Q = \frac{1}{2m} \sum_{i,j} \left( \mathbf{A}_{ij} - \frac{\hat{k}_i \hat{k}_j}{2m} \right) \delta(c_i, c_j)
\end{equation}
where \(Q\) is the modularity score, \(m\) is the total number of edges in the network, \(\mathbf{A}_{ij}\) is the adjacency matrix entry, \(\hat{k}_i\) and \(\hat{k}_j\) are the degrees of nodes \(v_i\) and \(v_j\), and \(\delta(c_i, c_j)\) is 1 if nodes \(v_i\) and \(v_j\) are in the same community, 0 otherwise.
% \begin{itemize}
%     \item \(Q\) is the modularity score,            
%     \item \(m\) is the total number of edges in the network,
%     \item \(\mathbf{A}_{ij}\) is the adjacency matrix entry, 
%     \item \(\hat{k}_i\) and \(\hat{k}_j\) are the degrees of nodes \(v_i\) and \(v_j\),
%     \item \(\delta(c_i, c_j)\) 
%     % is the Kronecker delta (
%     is 1 if nodes \(v_i\) and \(v_j\) are in the same community, 0 otherwise.
% \end{itemize}

There are many ways to detect communities in graphs, depending on the assumptions about meaningful partitions. Approaches like normalized cut aim to minimize cross-community connectivity by using cut-based objectives~\cite{shi_normalized_2000}. Most of the commonly used methods aim to optimize modularity~\cite {blondel_fast_2008, newman_finding_2004, traag_louvain_2019, clauset_finding_2004}. 
Since maximizing modularity is an NP-hard problem~\cite{brandes_maximizing_2006}, a common approach is to relax the discrete optimization into a spectral formulation. Specifically, the modularity function $Q$ can be rewritten in matrix form~\cite{newman_finding_2006} as:
\begin{equation}
\label{eq:spectral_modularity}
\mathbf{Q} = \frac{1}{2m} \text{Tr}\left( \mathbf{C}^\top \mathbf{B} \mathbf{C} \right),
\end{equation}
where \( \mathbf{C} \in \{ 0,1 \}^{n \times k}\) is a community membership matrix, with $C_{ij} = 1$ if node $v_i$ belongs to community $C_j$ and 0 otherwise, \( \mathbf{B}=\mathbf{A} - \frac{\mathbf{\hat{k}} \mathbf{\hat{k}}^\top}{2m} \in \mathbb{R}^{n \times n}\) is the modularity matrix, with \( \mathbf{\hat{k}} \) as a degree vector, and \( \text{Tr}(\cdot) \) is the trace matrix operator.
% \begin{itemize}
%     \item \( \mathbf{C} \in \{ 0,1 \}^{n \times k}\) is a community membership matrix, where $C_{ij} = 1$ if node $v_i$ belongs to community $C_j$ and 0 otherwise,
%     \item \( \mathbf{B}=\mathbf{A} - \frac{\mathbf{\hat{k}} \mathbf{\hat{k}}^\top}{2m} \in \mathbb{R}^{n \times n}\) is the modularity matrix, with \( \mathbf{\hat{k}} \) as a degree vector,
%     \item \( \text{Tr}(\cdot) \) is the trace matrix operator.
% \end{itemize}

\subsection{Community Concealment}
\label{sec:concealment_measures}
% Brief explanation of the general form of community hiding, i.e. problem formulation, measuring hiding, 
\textit{Community concealment} refers to a modification of a graph such that a specific group of nodes (i.e., a community) is no longer detected as a single group by a community detection algorithm. A widely used baseline for community concealment is Disconnect Internally, Connect Externally (DICE)~\cite{waniek_hiding_2018}, which randomly rewires intra-community edges to outside nodes under a fixed perturbation budget $b$. $b$ defines the total number of edge modifications (i.e., deletions and additions) applied to the graph.

Several metrics exist to assess a group's concealment success~\cite{fionda_community_2018, chang_community_2024}, but the most commonly used ones are derived by Waniek et al.~\cite{waniek_hiding_2018}.
% To measure the degree to which a group of nodes is successful at concealing, there are several metrics~\cite{fionda_community_2018, chang_community_2024}, but the traditionally used ones are derived by Waniek et al.~\cite{waniek_hiding_2018}. 
Assuming there is a community $C^\star$, or a target community, aimed to be concealed within other communities in $G$, then $M_1$ is defined by:
% $M_1(C^\star, C)$, where $C^\star$ is a target community (aimed to be concealed) within communities in $C$, is defined by:
\begin{equation}
M_1(C^\star, \mathcal{C})= \frac{|\{C_i\in \mathcal{C} : C_i \cap C^\star \neq \emptyset\}| - 1}{\max(|\mathcal{C}| -1, 1) \times \max_{C_i \in C} (|C_i \cap C^\star|)}
\end{equation}
\noindent
Thus, $M_1$ quantifies how well the nodes of $C^\star$ are spread out across the communities in $G$. 

\noindent Similarly, $M_2$ is defined by:
\begin{equation}
M_2(C^\star, \mathcal{C}) = \sum_{\substack{C_i : C_i \cap C^\star \neq \emptyset}} \frac{|C_i \setminus C^\star|}{\max(n - |C^\star|, 1)}
\end{equation}
where $C_i\setminus C^\star$ denotes the set of nodes in community $C_i$ that are not members of $C^\star$. $M_2$ quantifies how well nodes of $C^\star$ are ``hidden in the crowd'' with other nodes  $V\setminus C^\star$.
Both metrics, $M_1$ and $M_2$, have values in the range $[0,1]$ where higher values indicate that the target community $C^\star$ is more effectively hidden, i.e., concealed from a community detection algorithm.

\subsection{GNNs}
% Brief explanation of the learning mechanics of GNN.
% A growing body of research aims to improve community detection by incorporating information on nodes' attributes and their connections~\cite{bhattacharya_community_2023, park_unsupervised_2020, tsitsulin_graph_2023, ying_hierarchical_2018, bianchi_spectral_2020}. These methods leverage both network structure and node features to improve community assignments. 
% In this context, GNNs integrate node attributes with graph topology and perform nonlinear feature aggregation to learn expressive node embeddings.
% GNNs have emerged as a powerful approach in this context, as they integrate node attributes with graph topology, performing nonlinear feature aggregation to learn expressive node embeddings.
GNN-based community detection leverages both graph topology and node 
features to learn node representations used for community assignment
~\cite{bhattacharya_community_2023, park_unsupervised_2020,
tsitsulin_graph_2023, ying_hierarchical_2018, bianchi_spectral_2020}.

Let $\mathbf{H}^{(l)} \in \mathbb{R}^{n \times d_l}$ denote the node-embedding, where the $i$-th row $\mathbf{H}^{(l)}_{i:}$ is the $d_l$-dimensional representation of the node $v_i$ at the layer $l$, and $\mathbf{H}^{(0)} = \mathbf{X}$, the original node feature matrix.
A generic GNN layer can be written in matrix form as:
\begin{equation}
\label{eq:gnn_generic}
    \mathbf{H}^{(l+1)} = \phi^{(l)}\!\left( \mathbf{H}^{(l)},\, \text{AGG}^{(l)}\!\big(\mathbf{A}, \mathbf{H}^{(l)}\big) \right),
\end{equation}
where $\text{AGG}^{(l)}(\mathbf{A}, \mathbf{H}^{(l)}) \in \mathbb{R}^{n \times d_l}$ is a neighborhood aggregation operator that mixes each node’s representation with those of its graph neighbors based on $\mathbf{A}$, and $\phi^{(l)}(\cdot)$ is a learnable update function applied row-wise to produce the next-layer representations $\mathbf{H}^{(l+1)}$.
After $L$ layers of updates, the final matrix $\mathbf{H}^{(L)}$ serves as the learned node embedding space. Different GNN architectures (e.g., GCN~\cite{kipf_semi-supervised_2017}, GraphSAGE~\cite{hamilton_inductive_2017}, GAT~\cite{velickovic_graph_2018}) mainly differ in the design of $\text{AGG}^{(l)}$ and $\phi^{(l)}$.

One widely used architecture is the GCN, which implements aggregation via a normalized adjacency matrix defined by:
\begin{equation}
    \tilde{\mathbf{A}} = \hat{\mathbf{D}}^{-\frac{1}{2}} (\mathbf{A} + \mathbf{I}_n)\, \hat{\mathbf{D}}^{-\frac{1}{2}},
\end{equation}
where $\mathbf{I}_n$ is the $n \times n$ identity matrix and $\hat{\mathbf{D}}$ is the diagonal degree matrix of $(\mathbf{A} + \mathbf{I}_n)$, i.e., $\hat{\mathbf{D}}_{ii} = \sum_j (\mathbf{A} + \mathbf{I}_n)_{ij}$. 
Using this normalized adjacency matrix, node features are updated through the following transformation:
\begin{equation}
\label{eq:gcn}
    \mathbf{H}^{(l+1)} = \sigma \!\left( \tilde{\mathbf{A}}\, \mathbf{H}^{(l)}\, \mathbf{W}^{(l)} \right),
\end{equation}
where $\mathbf{W}^{(l)} \in \mathbb{R}^{d_l \times d_{l+1}}$ is a trainable weight matrix and $\sigma(\cdot)$ is a nonlinear activation (e.g., ReLU). 

% One of the widely adopted GNNs for node representation is the graph convolutional network (GCN)~\cite{kipf_semi-supervised_2017}. In this node representation learning, the adjacency matrix $A$ is first normalized as follows:
% \begin{equation}
%     \tilde{A} = \hat{D}^{-\frac{1}{2}}(A + I_n)\hat{D}^{-\frac{1}{2}},
% \end{equation}
% where $I_n$ is an identity matrix and $\hat{D}$ is a diagonal matrix of $(A + I_n)$, i.e., $\hat{D_{ii}} = \sum_j (A + I_n)_{ij}$.

% Using this normalized adjacency matrix, node features are updated through the following transformation:
% \begin{equation}
% \label{eq:gcn}
%     H^{(l+1)} = \sigma\left( \tilde{A} H^{(l)} W^{(l)} \right),
% \end{equation}
% where $H^{l}$ is the node representation at layer $(l)$ (with $H^{0} = X$, original input of node features), $W^{(l)}$ is a trainable weight matrix, and $\sigma(\cdot)$ is a nonlinear activation function (e.g., ReLU). 

GCN learning is guided by a loss function $\mathcal{L}$ that updates the weight matrix $\mathbf{W}^{(l)}$ in each layer during training. During training, loss gradients are computed using backpropagation, and $\mathbf{W}^{(l)}$ is updated using gradient descent, i.e., $\mathbf{W}^{(l)} \leftarrow \mathbf{W}^{(l)} - \alpha \frac{\partial \mathcal{L}}{\partial \mathbf{W}^{(l)}}$, where $\alpha$ is the learning rate. 

% Learning in GCNs is driven by a loss function $ \mathcal{L}$ that guides how the weight matrix $\mathbf{W}^{(l)}$ in each layer is updated during training. During training, gradients of the loss with respect to the weights are computed via backpropagation, and each weight matrix $\mathbf{W}^{(l)}$ is updated using gradient descent, i.e., 
% $\mathbf{W}^{(l)} \leftarrow \mathbf{W}^{(l)} - \alpha \frac{\partial \mathcal{L}}{\partial \mathbf{W}^{(l)}}$,
% where $\alpha$ is the learning rate. 

In supervised learning settings, GCNs are typically trained to predict node labels by minimizing a classification loss (e.g., cross-entropy) on a labeled subset of nodes. In unsupervised settings where true labels are not available, GCNs are trained to optimize alternative objectives. For representation learning, this may involve a reconstruction-based loss~\cite{kipf_variational_2016}, while in clustering tasks, the objective often targets community quality measures such as modularity~\cite{tsitsulin_graph_2023} or normalized cut~\cite{bianchi_spectral_2020}. Once the node embeddings are learned, clustering is performed directly on these representations to infer community assignments.

% DMoN is a recent approach that optimizes modularity while learning features end-to-end, achieved state-of-the-art results with imporvements of over 40\% over other methods. 

\subsection{Interpretable ML}
% \label{appx:shap_background}
Our work aims to identify the characteristics of communities that affect their concealment. We use interpretable ML, particularly feature importance~\cite{molnar_interpretable_2020}, to find these characteristics.

In \textit{feature importance} methods, input variables of a model (e.g., regression) are scored to assess their contribution to the prediction output. Popular methods such as mean decrease impurity (MDI)~\cite{louppe_understanding_2014} and permutation importance~\cite{breiman_random_2001}, can be biased toward high-cardinality features~\cite{strobl_bias_2007} or yield misleading results when features are highly correlated~\cite{nicodemus_behaviour_2010}. 
SHAP (SHapley Additive exPlanations)~\cite{lundberg_unified_2017} ensures each feature's contribution is additive and consistent, theoretically grounding the explanation of model prediction based on Lloyd Shapley’s approach to cooperative games~\cite{shapley_17_1953}.
According to the method, the contribution of each feature $i$ is calculated by averaging its marginal effect across all possible subsets of other features. The prediction for $x$ is the sum of all $M$ features' contributions: $f(x) = \phi_0 + \sum_{i=1}^{M} \phi_i, $ where $\phi_0$ is the model prediction's expected value and $\phi_i$ is the contribution of the $i$-th feature.
In this study, we use SHAP to determine which network-level and statistical descriptors strongly influence community concealment.  While our framework supports other interpretable techniques like LIME~\cite{ribeiro_why_2016}, we report only SHAP results for robustness and consistency~\cite{strumbelj_explaining_2014, samek_towards_2019}.

\section{Methods}
\label{sec:methods}

% Include a map of the experiments with each block, graphical for the methods we used.
%  Overview:
% 1. Synthetic Network Generation
% 2. Community Detection
% 3. Community Hiding Attack
% 4. Evaluation
% 5. Feature Importance Analysis

% \subsection{Experimental Design}
% explain the LFR and parameters, especially mu and minimal community  size

\subsection{LFR Benchmark}
\label{sec:lfr}
We conduct our experiments on synthetically generated networks, where we can precisely control the edge density both within and between communities. To simulate realistic community structures, we use the Lancichinetti–Fortunato–Radicchi (LFR) benchmark generator~\cite{lancichinetti_benchmark_2008}, a widely adopted method that produces graphs with ground-truth community labels. LFR is designed to generate graphs closely resembling properties observed in real-world networks, such as power-law degree distributions and power-law community sizes~\cite{fortunato_community_2010, lancichinetti_community_2009, yang_comparative_2016}.

% The degree distribution is controlled by the exponent parameter $\alpha$, while $\langle k \rangle$ is the average degree. 
Nodes are assigned degrees based on a power-law distribution with exponent $\alpha$, constrained by an average degree $\langle \hat{k} \rangle$ and a maximum degree $\hat{k}_{max}$. Nodes are assigned edges so that $1 - \mu$ of their connections are within their community and a fraction $\mu$ connects to nodes outside the community. 
When $\mu \rightarrow 0$, most edges are internal (strong, well-separated communities), while when $\mu \rightarrow 1$, most edges are external (indistinguishable communities). At approximately $\mu \approx 0.5$, nodes have equal external and internal edges, indicating the boundary where communities are no longer defined in the strong sense~\cite{lancichinetti_benchmark_2008}.
Power-law distribution determines community sizes as well, with $\beta$ as the exponent and $s_{\text{min}}$ as the minimum size, constrained by $s_{\text{min}}> \hat{k}_{min}$ and $s_{\text{max}}> \hat{k}_{max}$.

In our experiments, we generate networks using the LFR benchmark model with $N=1,000$ and examine community concealment under controlled parameters, including mixing parameter $\mu$ and minimum community size $s_{\text{min}}$. The complete set of experiment parameters is listed in Table~\ref{tab:experiments_params}. 

\begin{table}[!htbp]
\caption{Experimental Configuration Parameters.}
\label{tab:experiments_params}
\centering
\small
\begin{tabularx}{\columnwidth}{p{0.3cm}p{4.2cm}Y}
\hline
\multicolumn{2}{l}{\textbf{Parameter}} & \textbf{Value} \\
\hline
\multicolumn{3}{c}{\textit{LFR parameters}} \\
$N$ & Number of nodes & 1{,}000 \\
$\hat{k}_{max}$ & Maximum degree & $0.1N$ \\
$\langle \hat{k} \rangle$ & Average degree & 25 \\
$s_{\text{min}}$ & Minimum community size & 10, 30, 60 \\
$\alpha$ & Degree distribution exponent & $-2$ \\
$\beta$ & Community size distribution exponent & $-1.1$ \\
$\mu$ & Mixing parameter & 0.01, 0.1, 0.2, \dots, 0.5 \\
$\sigma_c$ & Feature centroid separation & 0.01, 0.1, 0.5, 1, 2, 5 \\
\hline
\multicolumn{3}{c}{\textit{DICE parameters}} \\
$\beta_b$ & Perturbation budget (\% of $|E_{\text{intra}}|$) & 0.01, 0.05, 0.1, \dots, 1 \\
$p$ & Deletion ratio parameter & 0, 0.25, 0.5, 0.75, 1 \\
\hline
\multicolumn{3}{c}{\textit{Experimental setup}} \\
$r$ & Number of realizations & 50 \\
\hline
\end{tabularx}
\end{table}

\subsection{Feature Generation for LFR}
\label{sec:features_LFR}
% explain how we added features based on the paper from Tsitsulin before DMoN
To simulate a graph learning scenario for evaluating community concealment methods, we extend the synthetic LFR benchmark by generating node features. Since LFR does not provide node features by default, we generate synthetic features for LFR using a method adapted from Tsitsulin et al.~\cite{tsitsulin_synthetic_2022} originally proposed for the Stochastic Block Model (SBM~\cite{snijders_estimation_1997}).

Assuming that LFR produces $k$ ground-truth communities denoted by ${C^{\text{LFR}}_1, C^{\text{LFR}}_2, \ldots, C^{\text{LFR}}_k}$, where each $C^{\text{LFR}}_i \subseteq V$ and $C^{\text{LFR}}_i \cap C^{\text{LFR}}_j = \emptyset$ for $i \ne j$. 
To generate aligned node features, we construct $k$ feature clusters $\mathcal{C} = \{C^f_1, \ldots, C^f_k\}$, such that each node $v_j$ has a feature vector $\mathbf{x}_j \in \mathbb{R}^d$ from a distribution associated with the feature cluster corresponding to its community. Formally:
\[
v_j \in C^{\text{LFR}}_i \quad \Rightarrow \quad \mathbf{x}_j \sim \mathcal{N}(\mu^f_i, \sigma^2 \mathbf{I}_{d \times d}) \quad \text{and} \quad \mathbf{x}_j \in C^f_i,
\]
where $\mu^f_i$ is the centroid of the feature cluster $C^f_i$. This setup ensures that the node community assignments align with feature clusters, i.e., $\mathcal{C}=\{C^f_1,\ldots,C^f_k\} \quad \text{with } C_i^{\text{LFR}} \equiv C^f_i$.

For each feature cluster $C^f_i$, we sample a centroid $\mu^f_i \in \mathbb{R}^d$ from a $d$-dimensional multivariate Gaussian distribution with zero mean and standard deviation $\sigma_c^2$:
\begin{equation}
    \mu^f_i\sim \mathcal{N}(0,\sigma_c^2 \cdot \mathbf{I}_{d\times d}),
\end{equation}
where $\sigma_c$ controls the dispersion between the centroids of clusters. We default to $d = 32$, as in other studies~\cite{tsitsulin_synthetic_2022, goel_community_2026}.

Then, for each node $v_j \in C^{\text{LFR}}_i$, we generate a feature vector $\mathbf{x}_j$ by sampling from another $d$-dimensional multivariate Gaussian distribution centered at the corresponding centroid $\mu^f_i$
\begin{equation}
    \mathbf{x}_j \sim \mathcal{N}(\mu^f_i,\sigma^2 \cdot \mathbf{I}_{d\times d}),
\end{equation}
where $\sigma^2 = 1$ is a fixed intra-cluster variance.

Note that $\sigma_c$ controls the separation between feature cluster centroids. A higher $\sigma_c$ means centroids are farther apart, making clusters more separable, while lower values lead to overlapping clusters. 
When $\sigma_c$ is much smaller than $\sigma$ (which we normalize by setting $\sigma = 1$), the feature clusters become highly overlapping, introducing noise that makes cluster boundaries harder to distinguish in feature space, and thus easier to conceal from a community detection algorithm. 

This synthetic feature generation process enables us to evaluate how well GNN-based community detection and subsequent community concealment methods perform under varying levels of feature noise.

\subsection{Baseline: DICE}
A widely used and simple method for community concealment is DICE (Disconnect Internally, Connect Externally)~\cite{waniek_hiding_2018}. The major advantage of DICE is that it does not require global knowledge of the entire graph. Instead, it relies on a modularity-inspired heuristic: reduce the internal connectivity of the target community while increasing its external connectivity to the rest of the network.

The algorithm is stochastic and follows a two-step approach: (1) edge deletion inside the community and (2) edge addition to external nodes. 
Thus, given a target community \( C^\star \subseteq G \); a perturbation budget \( b \in \mathbb{N} \), i.e., the total number of edge modifications; and a proportion parameter \( p \in [0, 1] \) that defines the fraction of the budget allocated to edge deletions (with the remaining $1- p$ fraction assigned to edge additions), the algorithm proceeds as in Algorithm \ref{alg:dice}:
% \begin{itemize}
%     \item a target community \( C^\star \subseteq G \),
%     \item a total perturbation budget \( b \in \mathbb{N} \) representing the number of edge modifications (deletions plus additions), and
%     \item a proportion parameter \( p \in [0, 1] \) that defines the fraction of the budget allocated to edge deletions (with the remaining $1- p$ fraction assigned to edge additions),
% \end{itemize}
% the algorithm proceeds as in Algorithm \ref{alg:dice}: 

\begin{algorithm}[H]
\caption{DICE}
\label{alg:dice}
\begin{algorithmic}
    \STATE \textbf{Input:} $G = (V, E)$, $C^\star \subseteq V$, $b \in \mathbb{N}$, $p \in [0, 1]$
    \STATE \textbf{Output:} Perturbed graph $G' = (V, E')$

    \STATE $G' \gets G$; $b_{\text{del}} \gets \lfloor b \cdot p \rfloor$; $b_{\text{add}} \gets b - b_{\text{del}}$
    \vspace{0.2em}
    \STATE \textit{1. Intra-community edge deletion}
    \STATE $E_{\text{intra}} \gets \{ (u, v) \in E \mid u \in C^\star, v \in C^\star \}$
    \STATE Randomly sample $E_{\text{del}} \subseteq E_{\text{intra}}$ such that $|E_{\text{del}}| = \min(b_{\text{del}},\ |E_{\text{intra}}|)$
    \STATE Remove edges $E_{\text{del}}$ from $G'$
    
    \vspace{0.3em}
    \STATE \textit{2. External edge addition}
    \STATE $V_{\text{ext}} \gets V \setminus C^\star$
    \STATE $\mathcal{E}_{\text{cand}} \gets \{ (u, v) \in C^\star \times V_{\text{ext}} \mid (u, v) \notin E \}$
    \STATE Randomly sample $E_{\text{add}} \subseteq \mathcal{E}_{\text{cand}}$, such that $|E_{\text{add}}| = \min(b_{\text{add}}, |\mathcal{E}_{\text{cand}}|)$
    \STATE Add edges $E_{\text{add}}$ to $G'$
    \vspace{0.2em}
    \STATE $E' \gets (E \setminus E_{\text{del}}) \cup E_{\text{add}}$
    \RETURN $G' = (V, E')$
\end{algorithmic}
\end{algorithm}

\noindent where:
\begin{itemize}
    \item $b_{\text{del}}, b_{\text{add}} \in \mathbb{N}$ are the number of intra-community edges to delete and external edges to add, respectively,
    \item $E_{\text{intra}} \subseteq E$ is the set of edges within the target community $C^\star$,
    \item $V_{\text{ext}} = V \setminus C^\star$ is the set of nodes outside of the target community,
    \item $\mathcal{E}_{\text{cand}} =\{ (u, v) \in C^\star \times V_{\text{ext}} \mid (u, v) \notin E \}$ is the set of all possible cross-community edges that can be added (i.e., edges not yet present in $G$), and
    \item $E_{\text{del}} \subseteq E_{\text{intra}}$ and $E_{\text{add}} \subseteq \mathcal{E}_{\text{cand}}$ are randomly sampled sets of edges to delete and add, respectively,
    \item the final output is a perturbed graph $G' = (V, E')$ with edge set $E' = (E \setminus E_{\text{del}}) \cup E_{\text{add}}$.
\end{itemize}
In this work, DICE serves as the reference for a structure-only community concealment method. Because it perturbs only graph topology while leaving node attributes unchanged, it provides a controlled baseline for evaluating whether incorporating feature information improves community concealment under GNN-based inference.

\subsection{GNN-based Community Detection: DMoN}
We apply Deep Modularity Networks (DMoN)~\cite{tsitsulin_graph_2023}, a GCN-based unsupervised clustering method, to detect communities in the perturbed and original graphs. DMoN has achieved top performance, generating clusters that strongly correlate with ground truth labels~\cite{tsitsulin_graph_2023}, and demonstrated significant robustness against targeted and adversarial perturbations~\cite{goel_community_2026} when compared to other state-of-the-art methods such as DiffPool~\cite{ying_hierarchical_2018} and MinCutPool~\cite{bianchi_spectral_2020}.

DMoN learns soft cluster assignments by optimizing a modularity-based objective end-to-end. The method follows two main steps: (1) node representation learning via graph convolutions, and (2) soft cluster assignments computed through differentiable pooling.
DMoN replaces the self-loop mechanism in GCN with a learnable skip connection to better preserve raw node features. The forward pass in the encoder is defined as:
\begin{equation}
    \mathbf{H}^{(l+1)} = \text{SeLU}\left( \hat{\mathbf{A}} \mathbf{H}^{(l)} \mathbf{W} + \mathbf{X} \mathbf{W}_{skip} \right),
\end{equation}
where $\mathbf{W}_{skip} \in \mathbb{R}^{d \times d}$ is a learnable skip connection matrix applied to the original features $\mathbf{X}$,  $\hat{\mathbf{A}} = \mathbf{D}^{-\frac{1}{2}}\mathbf{A} \mathbf{D}^{-\frac{1}{2}}$ is a normalized adjacency matrix without self-loops, and SeLU is the Scaled Exponential Linear Unit activation~\cite{klambauer_self-normalizing_2017}. 
Given the final layer node representations $\mathbf{H} \in \mathbb{R}^{n \times d}$ from the modified GCN, the cluster assignment matrix is $\mathbf{C} = \text{softmax}(\text{GCN}(\hat{\mathbf{A}}, \mathbf{X}))$, where  $\mathbf{C} \in \mathbb{R}^{n \times k}$ and 
% \begin{equation}
% \label{eq:softmax}
% \mathbf{C} = \text{softmax}(\text{GCN}(\hat{\mathbf{A}}, \mathbf{X})),
% \end{equation}
$k$ is the predefined number of clusters. Each row in $\mathbf{C}$ gives the soft assignment of a node to clusters and is normalized to sum to 1.

DMoN optimizes spectral modularity (a relaxed form) with collapse regularization and learns soft assignments. The loss function is defined as follows:
% \begin{equation*}
% \mathcal{L}_{\text{DMoN}}(C,A) = \underbrace{ -\frac{1}{2m} \text{Tr}\left( C^\top B C \right)}_{\text{modularity}} + \underbrace{\frac{\sqrt{k}}{n} \left\| \sum_i C_i^\top \right\|_F - 1}_{\text{regularization}},
% \end{equation*}

% \begin{equation}
% \label{eq:dmon_loss}
% \begin{split}
% \mathcal{L}_{\text{DMoN}}(\mathbf{C},\mathbf{A}) &= \underbrace{ -\frac{1}{2m} \text{Tr}\left( \mathbf{C}^\top \mathbf{B} \mathbf{C} \right)}_{\text{modularity}} \\
% &\quad + \underbrace{\frac{\sqrt{k}}{n} \left\| \sum_i \mathbf{C}_i^\top \right\|_F - 1}_{\text{regularization}},
% \end{split}
% \end{equation}
{\small
\begin{equation*}
\label{eq:dmon_loss}
\begin{split}
\mathcal{L}_{\text{DMoN}}(\mathbf{C},\mathbf{A}) &= \underbrace{ -\frac{1}{2m} \text{Tr}\left( \mathbf{C}^\top \mathbf{B} \mathbf{C} \right)}_{\text{modularity}} + \underbrace{\frac{\sqrt{k}}{n} \left\| \sum_i \mathbf{C}_i^\top \right\|_F - 1}_{\text{regularization}},
\end{split}
\end{equation*}
}
where 
$\mathbf{B}=\mathbf{A} - \frac{\mathbf{\hat{k}} \mathbf{\hat{k}}^\top}{2m}$ is a modularity matrix with $\mathbf{\hat{k}} \in \mathbb{R}^n$ as the degree vector and $\left\|\right\|_F$ is a Frobenius norm. Note that $\frac{1}{2m} \text{Tr}\left( \mathbf{C}^\top \mathbf{B} \mathbf{C} \right)$ is a spectral relaxation of modularity (Eq. (\ref{eq:spectral_modularity})) with $\mathbf{C}$ as the soft cluster assignment matrix. 
The regularization constrains the optimization and prevents gradient-based methods from converging to trivial solutions (i.e., all nodes in one cluster)~\cite{tsitsulin_graph_2023}. 
This regularization term is normalized to lie within the range $[0, \sqrt{k}]$ and is 0 when clusters are perfectly balanced in size, and $\sqrt{k}$ when all nodes collapse into a single cluster.

Additionally, to avoid gradient descent becoming trapped in local optima of the highly non-convex objective function, the authors apply dropout~\cite{srivastava_dropout_2014} to GCN representations before the softmax in the computation of $\mathbf{C}$.

\subsection{Experimental Setup}
% We then assess the success of the community hiding using $M_1$ and $M_2$ metrics. 
We systematically vary the parameters of the graph, features, and attacks to evaluate how community concealment depends on different factors.

We vary the mixing parameter $\mu$ across six values \{0.01, 0.1, 0.2, 0.3, 0.4, 0.5\} to control the level of inter-community edge mixing. Additionally, we vary the feature noise parameter $\sigma_c$ across \{0.01, 0.1, 0.5, 1, 2, 5\}, and experiment with three different minimum community sizes in LFR: 10, 30, and 60, to explore the impact of community size distribution. These minimum community sizes yield 25, 18, and 12 communities, respectively.

We use DICE with different perturbation budgets $\beta_b $, each representing a different percentage of intra-community edges in the target community $C^\star$. For each experiment, we calculate $b = \lfloor \beta_b |E_{\text{intra}}| \rfloor$, where $\beta_b \in [0.01, 1]$ is varied in $0.05$ increments. This enables us to simulate perturbations from 1\% to 100\% of intra-community edges.
We also evaluate performance under different values of the deletion ratio parameter $p \in \{0, 0.25, 0.5, 0.75, 1\}$, controlling the balance between internal edge removals $b_{del}$ and external edge additions $b_{add}$.

% We also experiment with different minimum community sizes in the LFR benchmark graph generator, using values of 10, 30, and 60 to explore how size distributions affect the results. The DICE attack is applied with a range of perturbation budgets from 0.01 to 1.0 (in increments of 0.05), targeting intra-community edges.
% We perform the attack by targeting each community individually, recording the evaluation metrics $M_1$ and $M_2$ for each case. Each experiment with the attack is repeated 50 times to account for stochastic variation.
In each run, we apply DICE to each target community and record $M_1$ and $M_2$ for each case. Each experiment is repeated 50 times to ensure stochastic robustness. All experiment parameters are listed in Table~\ref{tab:experiments_params}.

% \subsection{Improved Version of DICE}

% \section{Results}
\section{Analysis of Concealment}
\label{sec:results}
% 5.1: shap, 5.2: dice, 5.3: intuition, 5.4: moddice: 5.5: empirical sets
% https://arxiv.org/pdf/2509.24662
In this section, we present results of our experiments and analysis of community concealment. 
In Section~\ref{sec:dice}, we report the results of community concealment with a baseline technique, DICE.
In Section~\ref{sec:shap}, we discuss the methods we use to analyze features that contribute to the concealment of a community from DMoN, an unsupervised GNN-based clustering. 
Based on the findings from Sections~\ref{sec:dice} and~\ref{sec:shap}, we then introduce a novel method (FCom-DICE) in Section~\ref{sec:fcomdice}. 
% FCom-DICE enhances DICE to achieve greater concealment from GNN-based clustering. 
% In Section~\ref{sec:fcomdice} we describe the method, and in Section~\ref{sec:evaluation} we report results in comparison with the baseline.

\subsection{DICE Performance}
\label{sec:dice}
% performance of DICE on different perturbations (baseline)
%     b effect (p=0.5)
%     delta vs mu
%     delta vs sigma_c
%     heatmaps

\begin{figure*}[!htb]
    \centering
    \includegraphics[width=\textwidth]{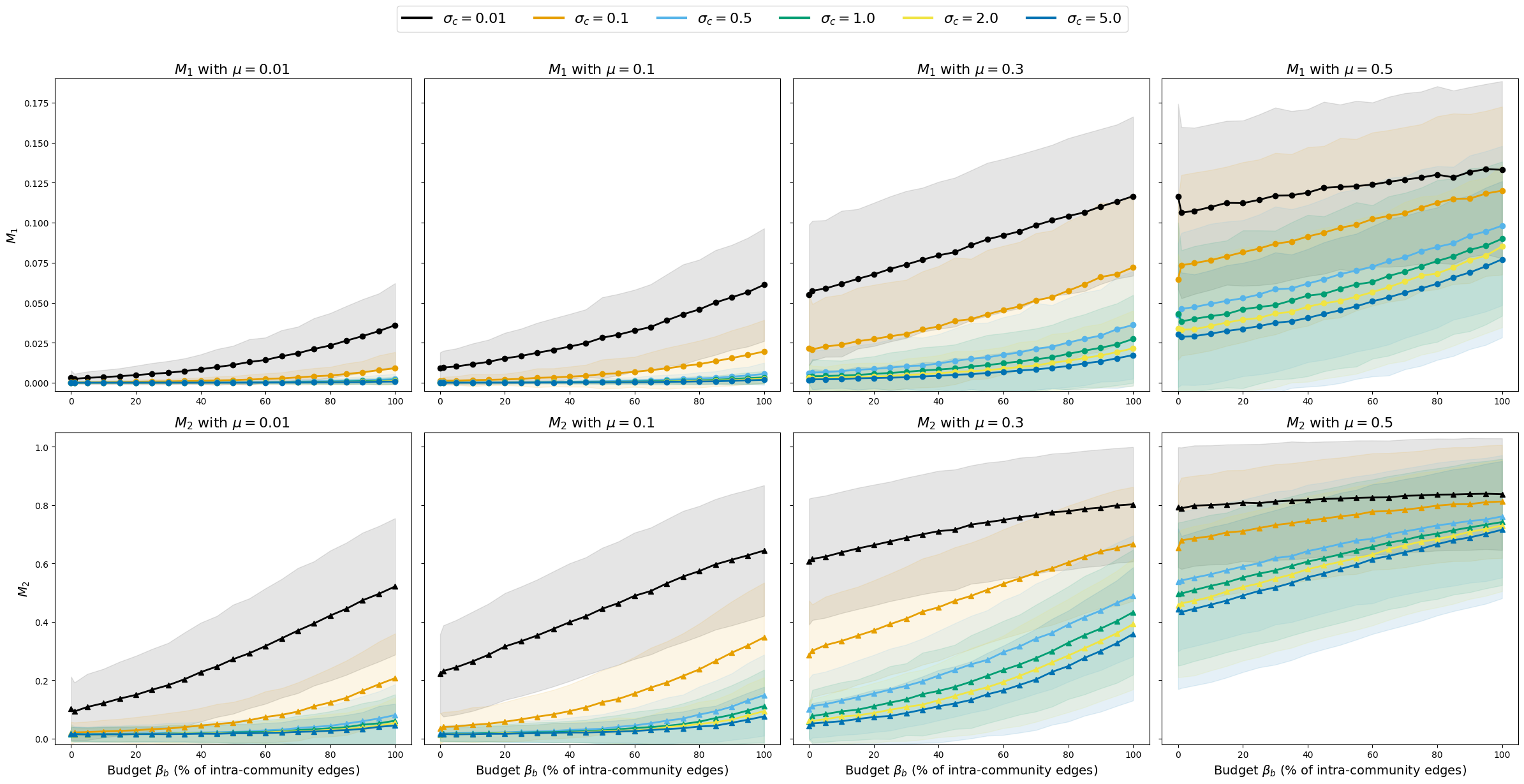}
    \vspace{-1em}
    \caption{Results of DICE performance with different $\sigma_c $, $\mu$, and perturbation budget $\beta_b$ with $p=0.5$ (50\% of the $b$ allocated to deletion vs adding edges) averaged over all realizations. Shaded bands around lines denote $\pm 1$ s.d. across all runs.}
    \label{fig:dice}
\end{figure*}

We start by investigating the performance of DICE with $p =0.5$ (the default for DICE), so half of the perturbation budget $b$ deletes intra-community edges in the target community, and the other half adds new external edges.
Figure~\ref{fig:dice} illustrates two metrics: $M_1$ (top row) and $M_2$ (bottom row), as a function of the perturbation budget $\beta_b$ (x-axis, representing the percentage of intra-community edges perturbed).
Columns denote the LFR mixing parameter $\mu \in \{0.01, 0.1, 0.3, 0.5\}$.
The colored curves represent the feature-space separation parameter $\sigma_c$ (a smaller $\sigma_c$ indicates increased overlap/noise), while the bands signify one standard deviation across all realizations of the experiments.
% Particularly, we use element centric similarity metric (ECS) to measure the accuracy of cluster assignment before and after perturbation is performed, compared to the ground truth provided by LFR. ECS computes similarity between two community assignments by comparing how nodes are grouped using PageRank affinity~\cite{}. ECS produces a score between 0 and 1, where 0 indicates no similarity and 1 indicates perfect matching of partitions. 

\begin{figure}[!htbp]
    \centering
    \includegraphics[width=\linewidth]{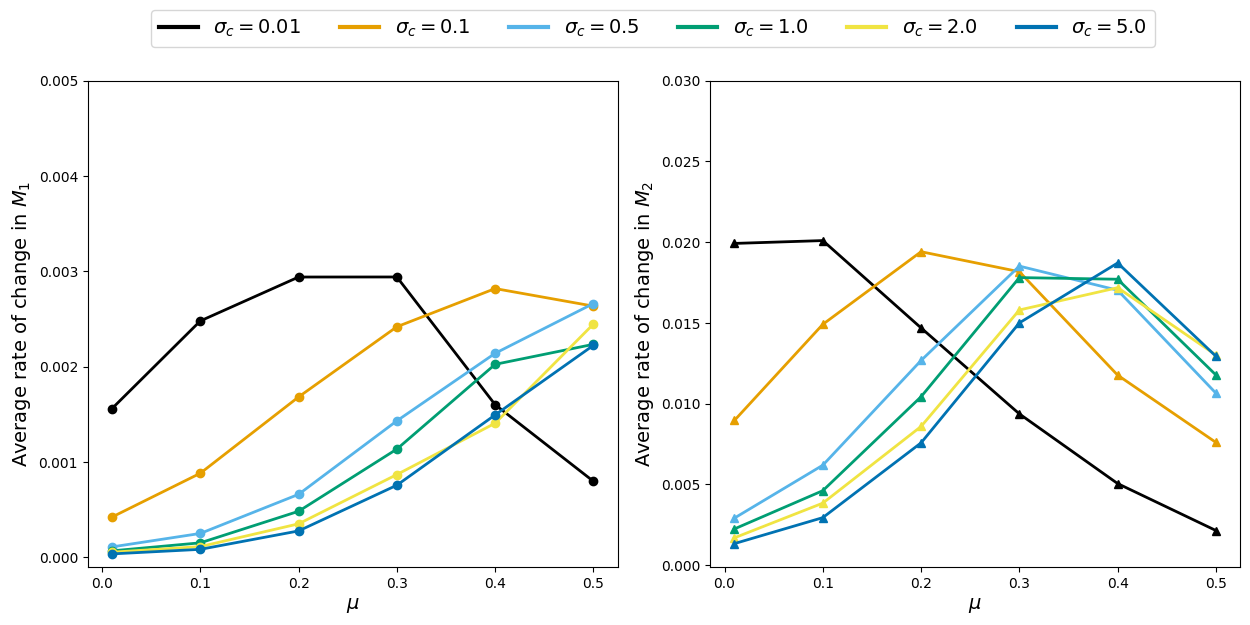}
    \vspace{-2em}
    \caption{Average rate of change of $M_1$ and $M_2$ vs. $\mu$ for each $\sigma_c$.}
    \vspace{-1em}
    \label{fig:avg_change}
\end{figure}

\begin{figure}[!htbp]
    \centering
    \includegraphics[width=\linewidth]{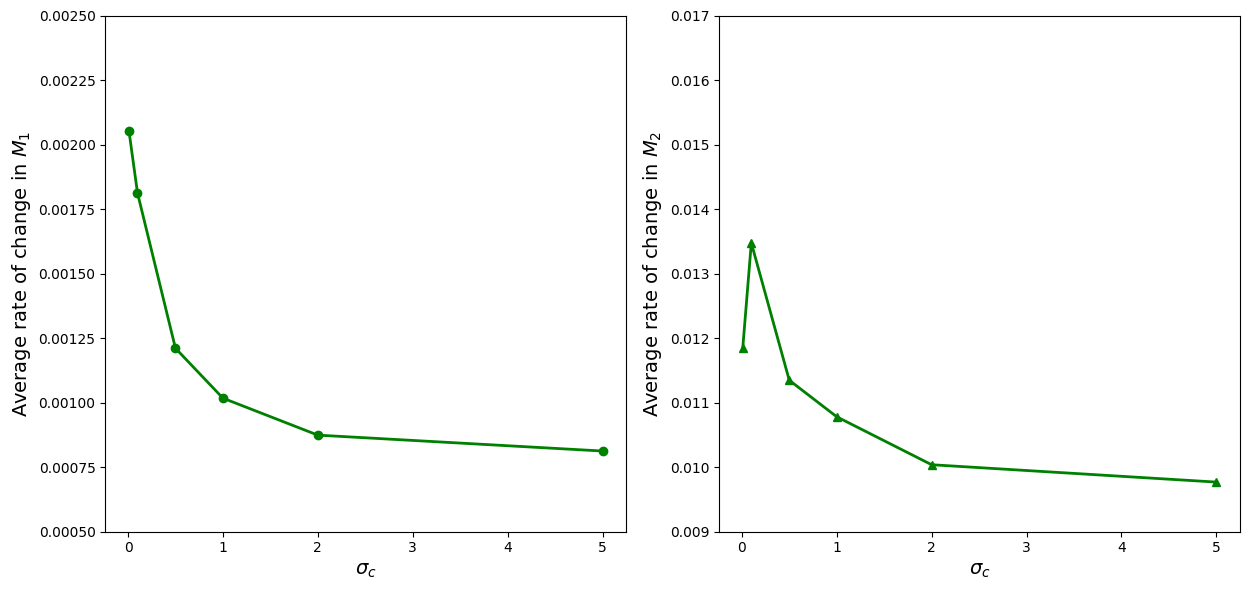}
    \vspace{-2em}
    \caption{Average rate of change of $M_1$ and $M_2$ vs. $\sigma_c$, averaged over $\mu$.}
    \vspace{-1em}
    \label{fig:avg_change_sigma}
\end{figure}

In all plots, increasing the perturbation budget $\beta_b$ consistently improves both metrics $M_1$ and $M_2$, indicating that greater perturbation enhances concealment.
Increasing the mixing parameter $\mu$ raises the curves for a constant budget $\beta_b$ and $\sigma_c$, making concealment easier as the ratio of inter-community to intra-community edges rises. 
A smaller $\sigma_c$ consistently leads to higher metric values due to overlapping features obscuring boundaries in the embedding space (as shown in Figure\ref{fig:dice}). This monotonic effect of $\sigma_c$ is statistically significant and consistent across all experimental configurations, as confirmed by a one-sided Jonckheere–Terpstra trend test~\cite{terpstra_asymptotic_1952,jonckheere_distribution-free_1954, ali_non-parametric_2015} aggregated via Stouffer’s method~\cite{stouffer1949american} ($Z=361.0$, $p \ll 10^{-6}$, one-sided).

\begin{figure*}[!htbp]
    \centering
    \includegraphics[width=0.7\textwidth]{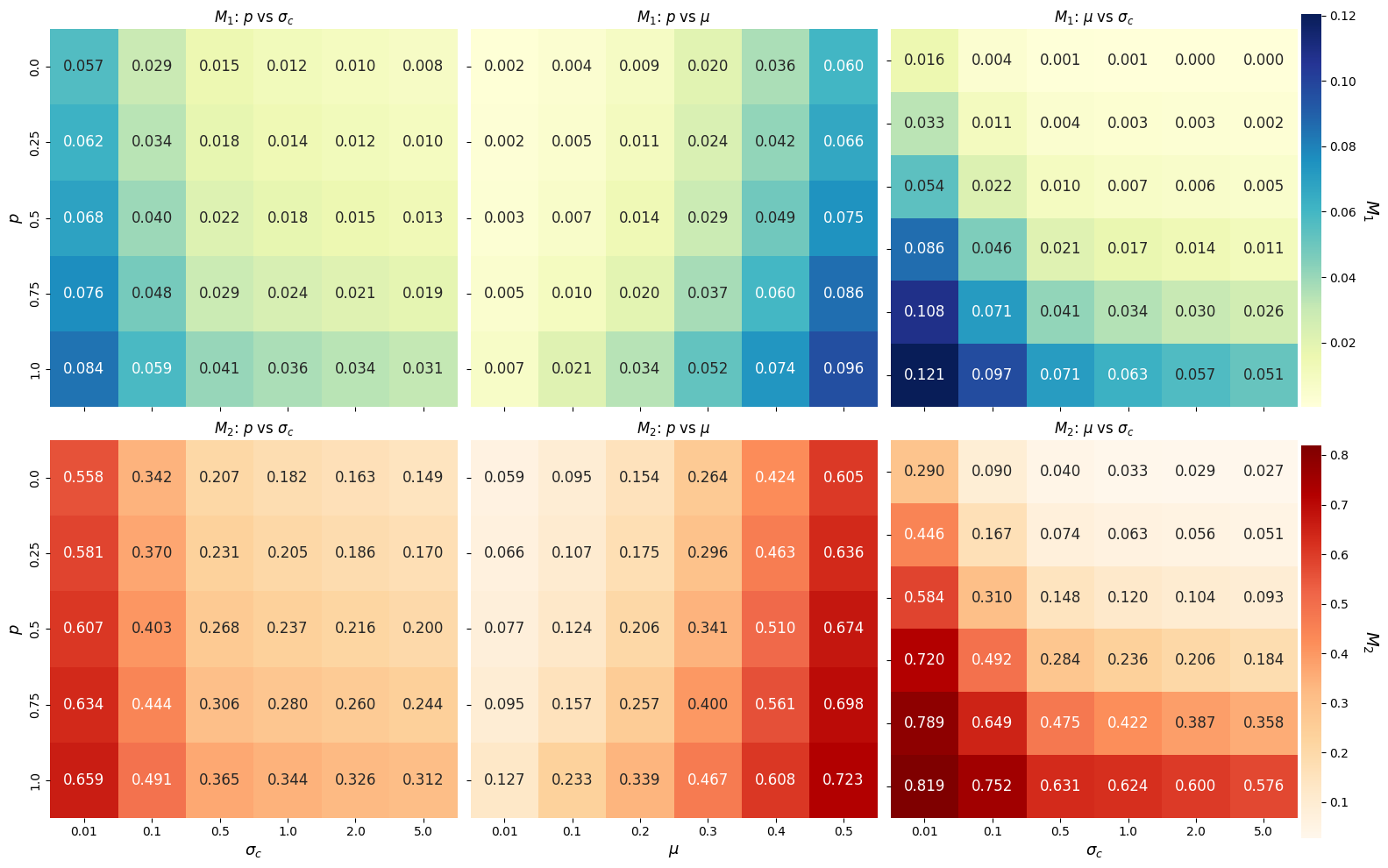}
    \caption{Heatmaps of $M_1$ and $M_2$ results of DICE performance with a combination of different $\sigma_c $, $\mu$, and $p$ averaged over all realizations, community labels and all $\beta_b$.}
    \label{fig:heatmap_dice_combo}
\end{figure*}

Note that $M_1$, although theoretically in the same range as $M_2$ (between 0 and 1) remains small in magnitude in all conditions (at most $\approx 0.175$ in our experiments), while still increasing gradually with $\beta_b$. Note that such a low magnitude of $M_1$ is typically observed in other studies as well~\cite{li_adversarial_2020}.
Separation between $\sigma_c$ curves is subtle at low $\mu$, however, becomes more apparent as $\mu$ increases. 
$M_2$, on the other hand, spans a much wider range and reacts more strongly to both $\mu$ and $\sigma_c$: for $\mu \in \{0.01, 0.1\}$ the increase with budget is roughly linear, while for $\mu \in \{0.3, 0.5\}$ the rise plateaus (approaching $M_2 \approx 0.8$ at $\mu=0.5$ with $\sigma_c=0.01$).  

% Figure~\ref{fig:avg_change} shows the average rate of change of $M_1$ and $M_2$ with respect to budget (i.e., $\delta M / \delta \beta_b$ ) against $\mu$.
Figure~\ref{fig:avg_change} shows the average finite rate of change of $M_1$ and $M_2$ with respect to the perturbation budget $\beta_b$, computed using mean finite differences ($\Delta M / \Delta \beta_b$) and plotted as a function of $\mu$.
For $M_1$ (left plot), when features are extremely overlapped ($\sigma_c =0.01$, black), the marginal gain peaks early (around $\mu\approx 0.2-0.3$) and drops as $\mu$ increases, indicating diminishing returns once there is enough structural noise. However, as features become more separable ($\sigma_c \geq 0.5$), the slope rises consistently with $\mu$, maxing at $\mu=0.5$. Thus, with more separable features, adding structural noise makes each increment of the perturbation budget more effective for $M_1$, however, with noisy features, added structural noise quickly inflates the effects, diminishing it after $\mu= 0.3$.
For $M_2$ (right plot), the pattern is similar. In Appendix~\ref{appx:dice_details} we show more details on this. 

The rate of change with highly noisy features ($\sigma_c = 0.01$) reaches its maximum at low $\mu$ and subsequently declines as $\mu$ increases, indicating diminishing returns thereafter. As $\sigma_c$ increases, creating distinct separation in feature space, the peak shifts to higher values of $\mu$; for instance, at $\sigma_c = 5$, the maximum rate of change occurs at $\mu = 0.4$.

% Figure~\ref{fig:avg_change_mu} illustrates the isolated impact of structural noise, depicting the average rate of change relative to budget as a function of $\mu$, averaged across all $\sigma_c$. As $\mu$ increases, the average rate of change in $M_1$ also increases, while the average rate of change in $M_2$ rises to a maximum at $\mu = 0.3$ before experiencing a slight decline thereafter. 
We plot the average rate of change vs. $\sigma_c$ and average over $\mu$ to isolate the impact of feature noise (Figure~\ref{fig:avg_change_sigma}). From Figure~\ref{fig:avg_change_sigma}, we can see that the average rate of change in $M_1$ decreases as $\sigma_c$ grows, as the more separable the features (larger $\sigma_c$), the less impact each unit of budget $\beta_b$ has. For $M_2$ the pattern is similar but with a small bump at $\sigma_c = 0.1$ before the overall decline. 
This could be explained by Figure~\ref{fig:avg_change}, where a very small $\sigma_c=0.01$ yields a high rate only at a small $\mu$ and then declines linearly, so its average over $\mu$ is pulled down. 

% \begin{figure}[!hbt]
%     \centering
%     \includegraphics[width=\linewidth]{./figures/avg change over all sigma.png}
%     \caption{Average rate of change of $M_1$ and $M_2$ vs. $\mu$, averaged over $\sigma_c$.}
%     \label{fig:avg_change_mu}
% \end{figure}

To further examine the performance of DICE, we vary $p \in [0, 0.25, 0.5, 0.75, 1]$, which controls how a perturbation budget $b$ is allocated between deleting intra-community edges and adding inter-community edges ($1-p$). 
In Figure~\ref{fig:heatmap_dice_combo}, each heatmap illustrates the metric value ($M_1$ at the top and $M_2$ at the bottom) averaged over all perturbation budgets $\beta_b$ for various combinations of $p$, $\sigma_c$, or $\mu$. 
For both, $M_1$ and $M_2$, increasing $p$ generally improves hidability, meaning deleting edges inside a community is more effective than adding. The strongest effect of $p$ is observed when $\mu$ is large and $\sigma_c$ is small, indicating structural and feature noise, respectively. 
Regardless of how the perturbation budget is divided, high $\mu$ and low $\sigma_c$ yield the largest value across all heatmaps. The rightmost heatmaps, which represent the average over $p$, indicate that $\mu$ and $\sigma_c$ are the predominant factors influencing the outcome of hidability, as values are the highest.

DICE's overall efficacy increases with the amount of community structure perturbation (budget $\beta_b$), the proportion of that perturbation devoted to intra-community edge deletion ($p$), the network's existing structural noise ($\mu$), and the overlapping of the node features ($\sigma_c$). 
Concealment is most difficult when communities are structurally well-separated and features are clean, and easiest when both the network's topology and node features blur community boundaries.
When features already blur communities (small $\sigma_c$), additional structural mixing yields smaller incremental benefits as $\mu$ grows. 
When features are more separable (larger $\sigma_c)$ increasing $\mu$ initially makes DICE’s budget more impactful, but once the network is highly mixed, the marginal returns begin to diminish.

\subsection{Analysis of Concealment Factors}
\label{sec:shap}
\subsubsection{Methods of Analysis}
\label{sec:shap_details}

\begin{table*}[!htbp]
\begin{center}
\caption{Measures used for community concealment analysis.}
\label{tab:community_measures}
\begin{tabular}{llp{10cm}}
\hline
\textbf{Measure} & \textbf{Notation} & \textbf{Description} \\
\hline
Average centroid sq. distance & $\overline{d^2}(C^\star)$ & Average squared Euclidean distance from the community $C^\star$ centroid to all other community centroids based on node features $\mathbf{X}$.\\
Community size & $|C^\star|$ & Number of nodes in the community $C^\star$ \\
Inter/Intra-edge ratio & $|E_{\text{inter}}|/|E_{\text{intra}}|$ & Ratio of external to internal edges of $C^\star$ \\
Mean degree centrality & $\overline{\hat{k}}_u, \; u \in C^\star$ & Average degree centrality of nodes in $C^\star$\\
Community degree centrality & $\hat{k}_{C^\star}$ & Degree of $C^\star$ as a super-node, i.e., number of edges leaving $C^\star$ \\
Mean betweenness centrality & $\overline{\mathrm{betw}}(u), \; u \in C^\star$ & Average betweenness centrality of nodes in $C^\star$ (fraction of shortest paths that pass through a node~\cite{freeman_set_1977})\\
Community betweenness & $\mathrm{betw}(C^\star)$ & Betweenness of $C^\star$ treated as a super-node (fraction of shortest paths between other communities that pass through $C^\star$) \\
Mean closeness centrality & $\overline{\mathrm{clos}}(u), \; u \in C^\star$ & Average closeness of nodes in $C^\star$ (average shortest distance from a node to all other nodes~\cite{freeman_centrality_1978}) \\
Community closeness & $\mathrm{clos}(C^\star)$ & Closeness of $C^\star$ as a super-node\\
\hline
\end{tabular}
\\
\footnotesize{\textit{Note:} $C^\star$ denotes the target community of an experiment.}
\end{center}
\end{table*}

To understand which properties of communities (before any perturbations) make them less detectable (more concealed) by GNN-based community detection, we rely on SHAP feature importance analysis~\cite {lundberg_unified_2017}. 
% The background on interpretable ML, particularly feature importance analysis is provided in Appendix~\ref{appx:shap_background}. 
In our experiments, we calculate measures of each community's internal structure and network relationships. These metrics include node-level centrality statistics and community-level metrics achieved by treating a community as a single ''super-node`` \cite{stanley_compressing_2018}.
Refer to Table~\ref{tab:community_measures} for the complete list and definitions. 

In addition to structural measures, we calculate feature-space separability of the target community $C^\star$ using a distance descriptor $\overline{d^2}(C^\star)$ defined as: 
\begin{equation}
\overline{d^2}(C^\star) \;=\; \frac{1}{|\mathcal{C}|-1}\sum_{C_i\in \mathcal{C}\setminus \{C^\star\}} \bigl\|\mu_{C^\star}-\mu_{C_i}\bigr\|_2^{\,2},
\end{equation}
where $\mathcal{C}=\{C_1,\ldots,C_k\}$ denotes the set of communities (aligned across LFR and node features, i.e., $C_i^{\text{LFR}}\equiv C_i^f\equiv C_i$), 
and $\mu_{C_i} = \frac{1}{|C_i|}\sum_{v\in C_i} x_v$ is the centroid (average vector) of community $C_i$ in a node feature space $\mathbf{X}$.

We model $M_1$ and $M_2$ separately, using predictors from Table~\ref{tab:community_measures}, randomly dividing data into 80\% training and 20\% testing. On the test set, we fit a \texttt{RandomForestRegressor} with default parameters and report performance as $R^2$. After calculating SHAP values using \texttt{TreeExplainer} on the test set, we aggregate the absolute values across data samples to determine global importance and rank features accordingly. 
Note that linear regression was rejected due to violated model assumptions, and instead, Random Forest Regression was used to capture non-linear relationships without distributional constraints.
% Note that we initially considered linear regression for modeling $M_1$ and $M_2$, but diagnostic tests showed that key assumptions (linearity, normality of residuals, homoscedasticity) did not hold. 
% Therefore, we adopted Random Forest Regression, which does not rely on these assumptions and can capture non-linear relationships between features and hidability measures. 

\subsubsection{Important Factors for Concealment}
With Random Forest Regression, we obtain a moderate predictive performance for both metrics ($R^2$ = 0.55 for $M_1$ and $R^2$ = 0.69 for $M_2$). 
% Figure~\ref{fig:shap} illustrates the summary of the SHAP analysis (beeswarm plot), which offers a comprehensive overview of SHAP values ($M_1$ in Figure~\ref{fig:shap-m1}, $M_2$ in Figure~\ref{fig:shap-m2}) for the feature set (refer to Table~\ref{tab:community_measures}) arranged by their impact. 
We observe that for both metrics $|E_{\text{inter}}|/|E_{\text{intra}}|$ (inter/intra-edge ratio) is the most influential predictor, followed by $\overline{d^2}(C^\star)$, the average centroid distance to other communities in $\mathbf{X}$. These two features explain 86\% of the total importance for $M_1$ and 79\% for $M_2$ (see Table~\ref{tab:shap_importance_table}). 

\begin{table}[!htbp]
\centering
\caption{Normalized global SHAP feature importances for $M_1$ and $M_2$.}
\label{tab:shap_importance_table}
\begin{tabular}{lcc}
\hline
\textbf{Feature} & $M_1$ \textbf{Importance}& \textbf{$M_2$ Importance} \\
\hline
$|E_{\text{inter}}|/|E_{\text{intra}}|$                  & \textit{0.487 (0.487)} & \textit{0.452 (0.452)} \\
$\overline{d^2}(C^\star)$                                & \textit{0.373 (0.860}) & \textit{0.334 (0.787)} \\
$\overline{\mathrm{clos}}(u)$                            & 0.047 (0.907) & 0.124 (0.911) \\
$\mathrm{clos}(C^\star)$                                 & 0.025 (0.932) & 0.037 (0.948) \\
$\overline{\hat{k}}_u$                                         & 0.021 (0.953) & 0.015 (0.980) \\
$|C^\star|$                                              & 0.018 (0.971) & 0.018 (0.965) \\
$\overline{\mathrm{betw}}(u)$                            & 0.017 (0.988) & 0.012 (0.992) \\
$\mathrm{betw}(C^\star)$                                 & 0.010 (0.998) & 0.006 (0.998) \\
$\hat{k}_{C^\star}$                                            & 0.002 (1.000) & 0.002 (1.000) \\
\hline
\end{tabular}
\\
\footnotesize{\textit{Note:} The importance score in the parentheses is reported as a cumulative share.}
\end{table}

\begin{figure}[!htbp]
  \centering
  \subfloat[$M_1$ model ($R^2=0.55$).\label{fig:shap-m1}]{
    \includegraphics[width=0.92\columnwidth]{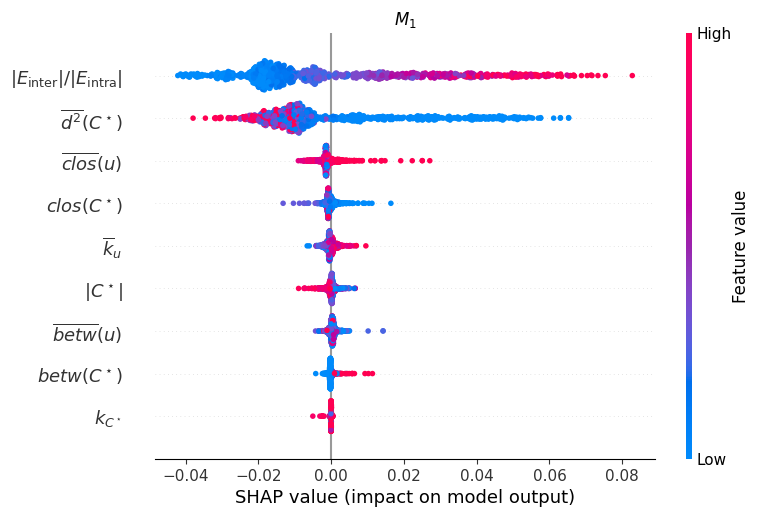}
  }
  \vspace{0.2em}
  \subfloat[$M_2$ model ($R^2=0.69$).\label{fig:shap-m2}]{
    \includegraphics[width=0.92\columnwidth]{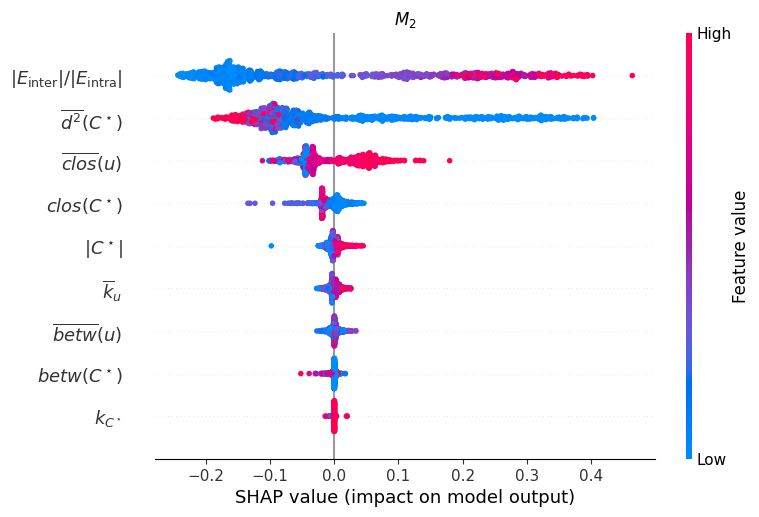}
  }
  \caption{SHAP summary plots for Random Forest Regression of community concealment.}
  \label{fig:shap}
\end{figure}
Figure~\ref{fig:shap} illustrates the summary of the SHAP analysis (beeswarm plot), which offers a comprehensive overview of SHAP values ($M_1$ in Figure~\ref{fig:shap-m1}, $M_2$ in Figure~\ref{fig:shap-m2}) for the feature set (refer to Table~\ref{tab:community_measures}) arranged by their impact. 
% Note that linear regression was rejected due to violated model assumptions and instead, Random Forest Regression was used to capture nonlinear relationships without distributional constraints.
Note that in both metrics ($M_1$ and $M_2$), higher values of $|E_{\text{inter}}|/|E_{\text{intra}}|$ (i.e., more external vs. internal edges) tend to increase the predicted value, and lower values of $\overline{d^2}(C^\star)$ (i.e., higher average similarity to other communities in $\textbf{X}$ space) also tend to push the predictions upward.
Thus, more external edges inside the target community and higher similarity of node features with other communities are important variables for hidability measured by $M_1$ and $M_2$. 
% A complete ranking with normalized (sum-to-1) importances and cumulative shares is presented in Table~\ref{tab:shap_importance_table}.

% Based on the findings from Sections~\ref{sec:dice} and~\ref{sec:shap}, we then introduce a novel method in Section~\ref{}.
% FCom-DICE, which enhances DICE to achieve greater concealment from GNN-based clustering. In Section~\ref{sec:fcomdice} we describe the method, and in Section~\ref{sec:evaluation} we report results in comparison with the baseline.
\section{Proposed Method: FCom-DICE}
% \subsection{Proposed Method: FCom-DICE}
\label{sec:fcomdice}
\subsection{Intuition}
Our experimental analysis of DICE and SHAP analysis shows that a target community's structural property is not the only factor in concealment from GNN-based community detection. Inter/intra-edge ratio and target community proximity based on node features are the best concealment predictors. Pre-existing structural and node feature-based noise helps DICE conceal a target community. Structure- and node-based noise improves DICE's ability to hide a target community. 
Thus, concealment improves when structural boundaries blur and target community nodes resemble those of other communities. 
This is consistent with how DMoN (and GNNs broadly) learns node representations and cluster assignments, aggregating node features from neighbors for each node.

Thus, our method's core idea is that perturbation should also be node feature-aware. Instead of adding edges arbitrarily (as in DICE), we propose that a target community's nodes attach to communities closest in node feature space to reinforce the node feature-based signal that pulls embeddings toward the same community in GNN learning.
Furthermore, we propose modifications to the node features of the perturbed nodes to enhance their similarity to the newly connected community. 

% \subsubsection{FCom-DICE: Feature-Community–guided DICE}
\subsection{FCom-DICE: Feature-Community-guided DICE}
Our proposed method, \textit{Feature-Community-guided DICE (FCom-DICE)}, is an adaptation of DICE that adheres to a two-step approach: (1) random edge deletion inside the target community $C^\star$ and (2) edge addition to an external node in the nearest community in feature space, with modification of the node features to the nearest community average.
Thus, given:
\begin{itemize}
    \item a target community \( C^\star \subseteq G \),
    \item a total perturbation budget \( b \in \mathbb{N} \) representing the number of edge modifications (deletions plus additions),
    \item a proportion parameter \( p \in [0, 1] \) that defines the fraction of the budget allocated to edge deletions,
    \item node features $\mathbf{X} = \{x_v \in \mathbb{R}^d : v \in V\}$,
    \item the community set $ \mathcal{C}=\{C_1,\ldots,C_k\} $ aligned across LFR and features
          $(C_i^{\text{LFR}}\equiv C_i^f\equiv C_i)$, where $ C_i=\{\,u\in V: c_u=i\,\} $,
    \item community labels $ \mathbf{c}=(c_1,\ldots,c_n)^\top $ with $c_u \in \{1,\ldots,k\}$, where $u \in V $,
    \item community feature centroids $ \displaystyle \mu_{C_i}=\frac{1}{|C_i|}\sum_{u \in C_i} x_{u} $,
    \item and a node–community similarity matrix $\mathbf{S}_{nc}\in\mathbb{R}^{|V|\times k}$ with
          entries $[\mathbf{S}_{nc}]_{u i} = -\bigl\|x_u - \mu_{C_i}\bigr\|_2^{\,2} $,
\end{itemize}
the algorithm proceeds as in Algorithm \ref{alg:fcom-dice}, 

\begin{algorithm}[!htbp]
\caption{FCom-DICE}
\label{alg:fcom-dice}
\begin{algorithmic}
    \STATE \textbf{Input:} $G = (V, E)$, $C^\star \subseteq V$, $b \in \mathbb{N}$, $p \in [0, 1]$
    \STATE \textbf{Output:} Perturbed graph $G' = (V, E')$ with modified node features $\mathbf{X'}$.

    \STATE $G' \gets G$; $b_{\text{del}} \gets \lfloor b \cdot p \rfloor$; $b_{\text{add}} \gets b - b_{\text{del}}$
    \vspace{0.2em}
    \STATE \textit{1. Intra-community edge deletion}
    \STATE $E_{\text{intra}} \gets \{ (u, v) \in E \mid u \in C^\star, v \in C^\star \}$
    \STATE Randomly pick $E_{\text{del}} \subseteq E_{\text{intra}} \mid |E_{\text{del}}| = \min(b_{\text{del}},\ |E_{\text{intra}}|)$
    \STATE Remove edges $E_{\text{del}}$ from $G'$: \quad $E' \gets E \setminus E_{\text{del}}$
    \vspace{0.3em}
    \STATE \textit{2. External edge addition with node feature modification}
    \STATE $V_{\text{ext}} \gets V \setminus C^\star$;\quad $\textit{added} \gets 0$
    \WHILE{$\textit{added} < b_{\text{add}}$}
        \STATE Randomly pick $u \in C^\star$
        \STATE $\mathcal K \gets \{i \in \{1,\ldots,k\}\setminus\{c_u\} \mid \exists v\in V_{\text{ext}} : c_v=i\ \wedge\ (u,v)\notin E' \}$
        \IF{$\mathcal K=\emptyset$} \STATE continue \ENDIF
        \STATE $i^\dagger \gets \arg\max_{i\in\mathcal K}\,[\mathbf S_{nc}]_{u i}$
        \STATE Randomly pick $v \in \{ w\in V_{\text{ext}} \mid c_w=i^\dagger,\ (u,w)\notin E' \}$ 
        \STATE Add edge $(u,v)$ to $G'$: \quad $E' \gets \big(E' \cup\ (u,v) \big)$
        \STATE \textit{Update feature vector for $u$:}
        \quad $x'_u \gets \mu_{C_{i^\dagger}}$ 
        \STATE $\textit{added} \gets \textit{added} + 1$
    \ENDWHILE
    \vspace{0.2em}
    \RETURN $G'=(V,E')$ with $\mathbf X'$
\end{algorithmic}
\end{algorithm}

\noindent where:
\begin{itemize}
    \item $b_{\text{del}}, b_{\text{add}} \in \mathbb{N}$ are the number of intra-community edges to delete and external edges to add, respectively,
    \item $E_{\text{intra}} \subseteq E$ is the set of edges within the target community $C^\star$,
    \item $G'$ is a perturbed graph with a modified edge set $E'$
    \item $V_{\text{ext}} = V \setminus C^\star$ is the set of nodes outside of the target community,
    \item $\mathcal{K}$ is a set of candidate destination communities for a node $u\in C^\star$ with a community label $c_u$,
    \item $i^\dagger$ is the community label of the \emph{most 
    feature-similar} feasible external community for $u$,
    \item  $\mu_{C_{i^\dagger}}$ is the feature centroid of the community $C_{i^\dagger}$,
    \item the final output is a perturbed graph $G' = (V, E')$ with $\mathbf X'$, which is $\mathbf X$ after all node feature edits.
\end{itemize}

In summary, FCom-DICE differs from DICE in the second step of the algorithm, i.e., edge addition. 
Rather than attaching to arbitrary external nodes, FCom-DICE selects, for a sampled $u\in C^\star$, the most feature-similar feasible destination community $i^\dagger \in \arg\max_{i\neq c_u} [\mathbf S_{nc}]_{u i}$, then chooses a non-neighbor $v\in C_{i^\dagger}$ and adds the edge $(u,v)$. 
With the edge addition, it modifies features of node $u$ to the destination community's centroid ($x'_u \gets \mu_{C_{i^\dagger}}$).
If no feasible destination exists for the sampled $u$, the iteration is skipped to the next sampled node.

The proposed community feature-based edge addition, coupled with feature modification, jointly reduces structural separability (more external than internal ties) and feature separability (greater similarity to a neighboring community). 

\section{Evaluation}
\label{sec:evaluation}
% proposed attack (mod-dice)
%     b effect (p=0.5)
%     delta vs mu
%     delta vs sigma_c
%     heatmaps
%     comparison with baseline (figure)
% Results of the suggested improved DICE and the measure of the improvement
% \subsubsection{Results on Featurized LFR}
\subsection{Results on Featurized LFR}
We first evaluate \textit{FCom-DICE} extensively against the baseline DICE using DMoN as the primary GNN-based community detector. DICE serves as the reference structure-only community concealment method, allowing us to evaluate whether incorporating feature-aware perturbations provides additional concealment benefits under GNN-based community inference. DMoN is used for the main evaluation because it directly optimizes the community detection quality function (i.e., modularity) and has been tested as the most robust GNN method for community detection~\cite{goel_community_2026}. 
We also evaluate the method against two additional GNN-based unsupervised clustering methods, DiffPool~\cite{ying_hierarchical_2018} and MinCut~\cite{bianchi_spectral_2020}. The two methods have been previously cross-compared with DMoN~\cite{tsitsulin_graph_2023, goel_community_2026}, making them appropriate for this paper. 
The results of experiments with these methods are presented in Section~\ref{sec:results_diffpool_mincut}, and details of each method and implementation settings are provided in Appendix~\ref{sec:methods_other_gnn}.

\begin{figure*}[!htbp]
    \centering
    \includegraphics[width=0.7\textwidth]{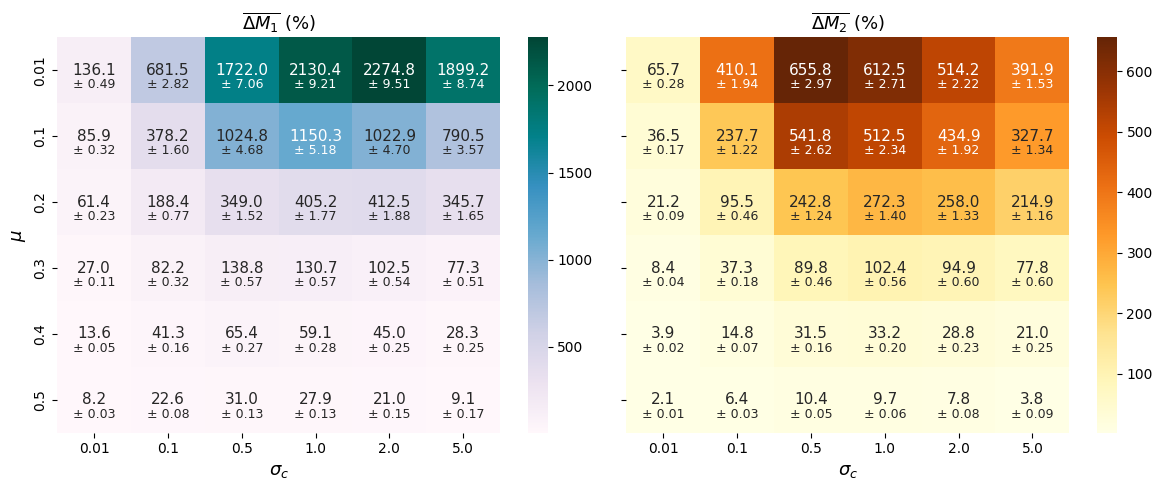}
    \caption{Average relative improvement (\% with $\pm 1$ s.d.) of FCom-DICE over baseline DICE averaged over all $\beta_b$ and all realizations using DMoN for community detection. The performance relative difference is shown for various combinations of $\mu$ and $\sigma_c$.
    }
    \label{fig:delta_fcom-dice}
\end{figure*}
% \begin{figure*}[htb!]
%     \centering
%     \includegraphics[width=\textwidth]{./figures/fcomdice per mu.png}
%     \caption{Results of FCom-DICE performance with different $\sigma_c $, $\mu$, and perturbation budget $\beta_b$ with $p=0.5$ averaged over all realizations. Shaded bands around lines denote $\pm 1$ s.d. across all runs.}
%     \label{fig:fcom-dice}
% \end{figure*}
% Our proposed method, FCom-DICE, yields a noticeable performance improvement over DICE (baseline). In Figure~\ref{fig:fcom-dice}, both $M_1$ (top row) and $M_2$ (bottom row) still increase with the perturbation budget $\beta_b$, and at the same time the overall trend of growing hidability with higher $\mu$ and lower $\sigma_c$ still holds. 
% However, with FCom-DICE, performance increases faster, particularly for larger $\mu$ (e.g., $\mu =0.3$ or $0.5$), reaching a plateau value only at perturbation budget $\beta_b \ 20\%$  , rather than gradually increasing across the entire budget range (as DICE in Figure~\ref {fig:dice}).
% For instance, at $\mu=0.5$ a low $\sigma_c$, $M_2$ rises to $\approx 0.8$ after less than 20\% of the budget, suggesting that the cap of the hiding capacity is achieved with a smaller perturbation compared to DICE.
% In general, FCom-DICE produces higher performance over DICE (baseline) for both metrics, $M_1$ and $M_2$, across different $\mu$ and $\sigma_c$. In Appendix~\ref{appx:fcom-dice_vs_dice}, we present a thorough comparison of the behavior of two methods over the perturbation budget $\beta_b$.

For the main evaluation we calculate $\overline{\Delta M_1}$ and $\overline{\Delta M_2}$, the average relative difference (in \%) for each $(\mu,\sigma_c)$ pair, averaged across all perturbation budgets $\beta_b \in \mathcal{B}$, all realizations $r \in \mathcal{R}$, and all target communities $C^\star \in \mathcal{C}^\star$ that are aimed to be concealed, and define it by:
\begin{equation}
\begin{split}
\overline{\Delta M_j}(\mu, \sigma_c) &=
\\
&\frac{1}{|\mathcal{B}|\,|\mathcal{R}|\,|\mathcal{C}|}
\sum_{\beta_b \in \mathcal{B}}
\sum_{r \in \mathcal{R}}
\sum_{C^\star \in \mathcal{C}}
\Delta M_j(\beta_b, r, C^\star; \mu, \sigma_c),
\end{split}
\label{eq:delta_M}
\end{equation}
where $\Delta M_j(\beta_b, r, C^\star;\mu,\sigma_c)$ is the percent relative difference between FCom-DICE and DICE when concealing the target community $C^\star$ under budget $\beta_b$ in realization $r$:
% where $\mathcal{B}$ is the set of tested perturbation budgets $\beta_b$, $\mathcal{R}$ is the set of realizations for each target community $C^\star\in \mathcal{C}$, and $\Delta M_j(\beta_b, r, C^\star; \mu, \sigma_c)$ is a relative difference in $M_j \in \{M_1, M_2\}$ between FCom-DICE and DICE concealing community $C^\star$ with $\beta_b$, $\mu$, $\sigma_c$ in realization $r$, defined as:
{\small
\begin{equation*}
    \Delta M_j(\beta_b, r, C^\star; \mu, \sigma_c) = 
    \left(
    \frac{M_j^{\text{FCom-DICE}} - M_j^{\text{DICE}}
    }{M_j^{\text{DICE}}}
    \right)
\times 100\%.
\end{equation*}
}
Here, $M_j^{\text{FCom-DICE}}$ and $M_j^{\text{DICE}}$ denote the achieved values of the metric $M_j \in \{M_1,M_2\}$ for FCom-DICE and DICE, respectively.

\subsubsection{DMoN}
\label{sec:results_dmon}
In general, FCom-DICE produces higher performance over DICE (baseline) for both metrics, $M_1$ and $M_2$, across different $\mu$ and $\sigma_c$. In Appendix~\ref{appx:fcom-dice_vs_dice}, we present a thorough comparison of the behavior of two methods over the perturbation budget $\beta_b$ when DMoN is used for community detection.

The heatmap in Figure~\ref{fig:delta_fcom-dice} illustrates $\overline{\Delta M_1}(\mu, \sigma_c)$ and $\overline{\Delta M_2}(\mu, \sigma_c)$, indicating the degree to which FCom-DICE improves upon the baseline DICE when DMoN is used for community detection. 
The improvement is largest when $\mu$ is small, i.e., when communities are structurally well separated and thus intrinsically harder to hide. At $\mu=0.01$, FCom-DICE yields the highest relative gain: $M_1$ exceeding 2,000\% change in $\sigma_c \in [1.0, 2.0]$, and $M_2$ exceeding 500\% in $\sigma_c \in [0.5,1.0, 2.0]$. 
However, as $\mu$ increases, the relative improvement diminishes but still shows at least an 8.2\% difference over DICE in $M_1$ and 2.1\% in $M_2$ when $\sigma_c = 0.01$ (when node features are overlapping the most). 
Intuitively, when $\sigma_c$ is very low and $\mu$ is high, it becomes difficult to distinguish clusters in both the node feature space and the topological structure. 
Consequently, improvements over DICE are minimal because DICE already performs well at concealing communities in this noisy regime. Therefore, incorporating feature-aware edge attachment and modifying node features to align with the average of the attaching community does not significantly enhance hidability, as the node features lack sufficient separability already.

The effect of $\sigma_c$, however, is not linear, but both, $\overline{\Delta M_1}(\mu, \sigma_c)$ and $\overline{\Delta M_2}(\mu, \sigma_c)$, peak at moderate $\sigma_c$. The largest relative gains tend to occur around $\approx 0.5-2.0$. 
A plausible explanation for this pattern might be provided by the GNN learning, particularly how DMoN learns clusters by optimizing modularity.  
Clusters are easily separable in node features $\mathbf{X}$ when $\sigma_c$ is large ($\sigma_c = 5$). As a result, when $\mu$ is low, the graph structure also reinforces this separation because the majority of edges stay within a community. 
FCom-DICE outperforms the baseline DICE, as random nodes selected inside the target community $C^\star$ are not only attached to other communities but also change their features to match the attached community. This simultaneously weakens $C^\star$ structurally and makes those nodes easier for DMoN to assign elsewhere.

However, when $\mu$ is high (and $\sigma_c$ is large), $C^\star$ already has many external edges relative to its internal edges. Structurally, this means communities are less separable and therefore make modularity-based clustering harder for DMoN even before any attack.
At the same time, high $\sigma_c$ makes communities highly distinct in the feature space. When DICE perturbs edges, leading to more noise in the structure, a GNN such as DMoN observes nodes from $C^\star$ densely connected to other communities, so during message passing, their features get pulled toward those neighboring communities.
Because the neighboring communities have clean, well-defined feature signals, these boundary nodes of $C^\star$ acquire embeddings that resemble the external communities rather than $C^\star$. As a result, when $\mu$ is high, baseline DICE already hides $C^\star$ quite effectively, which leaves less room for FCom-DICE to improve over it. 

Overall, FCom-DICE surpasses DICE, especially demonstrating significant relative improvement in networks with low $\mu$. It is important to note that the perturbations introduced by FCom-DICE do not alter the overall network community structure, as the community structure remains largely unchanged after the perturbation, even with significant perturbation budgets (refer to Appendix~\ref{appx:fcom-dice_ecs}).
In addition, compared to DICE, FCom-DICE requires only a perturbation budget of 20-30\% to reach the high plateau of hidability (see Appendix~\ref{appx:fcom-dice_vs_dice} for details). 

\subsubsection{DiffPool and MinCut}
\label{sec:results_diffpool_mincut}

To further evaluate the performance of FCom-DICE over DICE, we additionally compute the results using DiffPool and MinCut as the underlying GNN-based community detection models. Note that DMoN was reported to perform better and more stably than these two methods~\cite{tsitsulin_graph_2023, goel_community_2026}.
% Details of the DiffPool and MinCut methods are provided in Appendix~\ref{sec:methods_other_gnn}. 

Unlike the DMoN results, we do not report the average relative difference $\overline{\Delta M_1}$ and $\overline{\Delta M_2}$ for DiffPool and MinCut. In multiple noisy settings, both models tend to produce trivial or near-trivial cluster assignments, assigning all nodes to one very large community, or to only a few large communities. This behavior makes $\overline{\Delta M_1}$ and $\overline{\Delta M_2}$ difficult to compute and interpret. In particular, when the detected community spans almost the entire graph, the $M_1$ score is often $0$, as target-community nodes are not spread across other communities (see Section~\ref{sec:concealment_measures}). This makes the relative change $\overline{\Delta M_1}$ numerically unstable. At the same time, $M_2$ can be close to $1$, as the large detected community contains many non-target nodes. 
% We run DiffPool and MinCut using the parameters that experimentally produced the best partitioning compared to the ground truth. (see Appendix~\ref{sec}).

\begin{figure*}[!htbp]
    \centering
    \subfloat[DiffPool]{
        \includegraphics[width=0.5\textwidth]{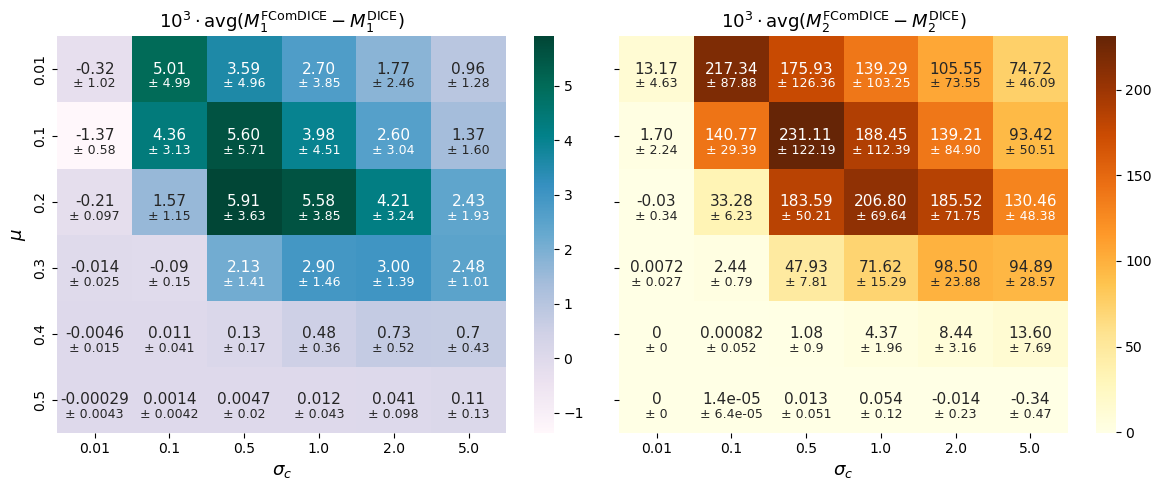}
        \label{fig:diffpool_abs_change}
        }
    \subfloat[MinCut]{
        \includegraphics[width=0.5\textwidth]{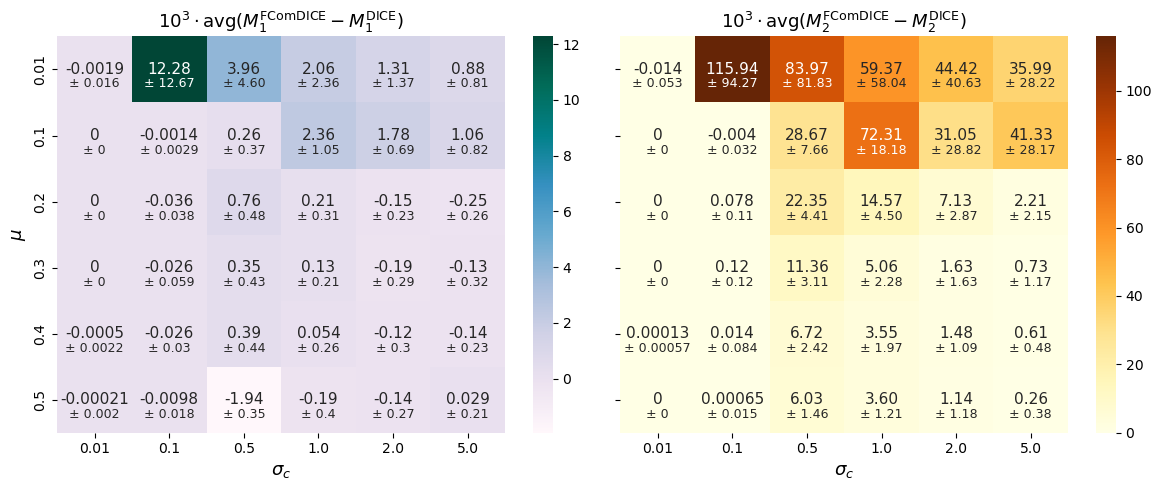}
        \label{fig:mincut_abs_change}
        }
    \caption{Scaled average difference (with $\pm 1$ s.d.) between FCom-DICE and baseline DICE averaged over all $\beta_b$ and all realizations using DiffPool and MinCut for community detection. The performance difference is shown for various combinations of $\mu$ and $\sigma_c$.
    }
    \vspace{-0.5em}
    \label{fig:diffpool_mincut_results}
\end{figure*}

Therefore, for DiffPool and MinCut, Figure~\ref{fig:diffpool_mincut_results} reports the scaled average difference between FCom-DICE and DICE:
\[
10^3 \cdot 
% \mathrm{avg}
\frac{1}{|\mathcal{B}|\,|\mathcal{R}|\,|\mathcal{C}|}
\sum_{\beta_b \in \mathcal{B}}
\sum_{r \in \mathcal{R}}
\sum_{C^\star \in \mathcal{C}}
(M_j^{\text{FCom-DICE}} - M_j^{\text{DICE}})
\]
where the average is computed over perturbation budgets, target communities, and realizations for each $\mu$ and $\sigma_c$ pair. Positive values indicate that FCom-DICE achieves higher hidability than DICE.

Overall, the results on DiffPool and MinCut confirm that FCom-DICE improves over DICE. For DiffPool, the improvement is more visible at lower $\mu$ and higher $\sigma_c$, with the strongest gains appearing at intermediate $\sigma_c$. For MinCut, the same general pattern holds, with improvements concentrated at $\mu=0.01$ and $\mu=0.1$. The improvements in $M_1$ are smaller and less stable, which is consistent with MinCut's tendency to form trivial clusters in noisy settings.

Although DiffPool and MinCut are less reliable than DMoN in recovering community structure~\cite{goel_community_2026}, FCom-DICE still generally increases the concealment of a community over DICE.

% \begin{equation}
% % \Delta M_j(\beta_b, r, C^\star;\mu,\sigma_c)
% % =
% \left(
% \frac{
% M_j^{\text{FCom}}(\beta_b,\mu,\sigma_c,r,C^\star)
% -
% M_j^{\text{DICE}}(\beta_b,\mu,\sigma_c,r,C^\star)
% }{
% M_j^{\text{DICE}}(\beta_b,\mu,\sigma_c,r,C^\star)
% }
% \right)
% \times 100\% .
% \label{eq:delta_single}
% \end{equation}

% \begin{equation}
% \begin{aligned}
% \overline{\Delta M_1}(\mu,\sigma_c)
% = 
% \frac{1}{|\mathcal{B}|\,|\mathcal{R}|\,|\mathcal{C}|}
% \sum_{\beta_b \in \mathcal{B}}
% \sum_{r \in \mathcal{R}}
% \sum_{C^\star \in \mathcal{C}}
% \\
% \cdot
% \frac{
%     M_1^{\text{FCom}}(\beta_b,\mu,\sigma_c,r,C^\star)
%     -
%     M_1^{\text{DICE}}(\beta_b,\mu,\sigma_c,r,C^\star)
% }{
%     M_1^{\text{DICE}}(\beta_b,\mu,\sigma_c,r,C^\star)
% }
% \times 100\% ,
% \end{aligned}
% \label{eq:delta_M1}
% \end{equation}

% \noindent\textbf{Results on Real Networks.}
\subsection{Results on Real Networks}

We evaluated the performance of FCom-DICE against the baseline DICE on three real networks: Wikipedia, Facebook and Bitcoin Transactions. 
The size, sparsity, attribute dimensionality, and label semantics of these datasets vary. The dataset statistics, community labels, and node features are detailed in Section~\ref{appx:real_datasets}, and evaluation results are presented in Section~\ref{sec:real_networks_results}
% Appendix~\ref{appx:real_datasets}.

\subsubsection{Experimental Details}
\label{appx:real_datasets}

% - Descrtiption of dataset, includiong table with stats.
% - Ground truth labels and their meaming:
%     - FB - ego netwokrs, and has signgletons. therefore we can't use this graound truth labels as assumed communities, as they are already hidden pretty nicely in the netwrok. 
%     - in Wikipedia labels are categories of the pages. 
% Mention that community labels are not 
%     - BTC has 3 classes ohnly as illegal/legal transaction and unknown, so we cannot use these classes as community labels. We sample only LCC

% appendix: Table with data descriotion, including number of community lables
% appendix: Method of creating labels (mention how true labels do not reflect the community using references and maybe actual results)
% appendix: Method on featurizing the real networks

We use real attributed networks provided by PyTorch Geometric. For Wikipedia and Facebook, we used \texttt{AttributedGraphDataset} class~\cite{yang_pane_2023}. For Bitcoin Transactions (BTC), we used the Elliptic Bitcoin dataset~\cite{weber_anti-money_2019} and limited analysis to the largest connected component to sample a sizable network of transactions. Details for each of the dataset is outlined in Table~\ref{tab:real_networks_stats}. 

\begin{table}[!htbp]
\centering
\caption{Statistics of real-world datasets used in evaluation.}
\label{tab:real_networks_stats}
\begin{tabular}{lrrr}
\hline
 & Wiki & Facebook & BTC$^\star$ \\
\hline
Nodes & 2,405 & 4,039 & 7,880 \\
Edges & 11,596 & 88,234 & 9,164 \\
Density & 0.0040 & 0.0108 & 0.0003 \\
Modularity & 0.505 & 0.482 & 0.217 \\
Original Features & 4,973 & 1,283 & 165 \\
Provided Labels & 17 & 151 & 3 \\
% Assortativity (with provided labels) & 0.564 & 0.535 & 0.499 \\
$|\mathcal{C}_{\text{consensus}}|$ & 58 & 16 & 39 \\
\hline
\end{tabular}
\\
\footnotesize{\textit{Note:} $|\mathcal{C}_{\text{cons}}|$ denotes the number of community labels obtained via consensus Louvain. \\
$^\star$The largest connected component (LCC) of the Bitcoin Transactions network was used.}
\end{table}

\noindent \textbf{Community Labels.}
The original node labels provided with the real-world datasets are not intended to represent ground-truth communities. Instead, they encode dataset-specific metadata whose semantics might not reflect a community or cluster in a graph. 
For Facebook, the graph is constructed from ego networks of users (nodes), and the provided labels correspond to ego-centric user attributes. These labels, which contain a lot of singleton classes, are therefore considered unsuitable for the community concealment downstream tasks covered in the paper. 
In Wikipedia's page-to-page hyperlink network, node labels correspond to topical categories assigned to pages (nodes). While semantically meaningful, these categories are broad and not defined by network connectivity and frequently span multiple structural communities.
For Bitcoin Transactions, only three labels are provided, indicating whether a transaction is classified as licit, illicit, or unknown. These classes reflect transactional status rather than structural organization and therefore cannot be interpreted as communities.

Therefore, we do not treat the provided labels as ground-truth communities and do not evaluate methods against the provided labels. Moreover, apart from the arguments described above, prior work has repeatedly shown that metadata labels in real networks often exhibit weak or inconsistent alignment with communities detected~\cite{hric_community_2014,peel_ground_2017}. 

Instead of relying on dataset-provided labels, we derive structural community assignments using consensus clustering based on the Louvain algorithm~\cite{blondel_fast_2008}, which is a widely used modularity-maximization method that is considered well-suited for large real-world networks. 
As Louvain is stochastic and may produce different partitions across runs, we obtain a stable community label using the consensus clustering method~\cite{lancichinetti_consensus_2012}.  Consensus clustering identifies pairs of nodes that consistently co-occur in the same community across repeated Louvain runs, constructs an agreement matrix encoding co-assignment frequencies, and retains only node pairs whose co-assignment frequency exceeds a threshold $\tau$, reclustering this agreement to produce the final consensus partition. 

In our experiments, we compute Louvain consensus clustering using 50 independent runs and set $\tau = 0.3$, consistent with reported high accuracy~\cite{lancichinetti_consensus_2012}. Note that consensus Louvain is used only to define a stable reference partition. 

% appendix: Table with data descriotion, including number of community lables
% appendix: Method of creating labels (mention how true labels do not reflect the community using references and maybe actual results)
% appendix: Method on featurizing the real networks

\subsubsection{Evaluation Results}
\label{sec:real_networks_results}

\begin{figure*}[!h]
    \centering
    \subfloat[DMoN]{
    \includegraphics[width=0.7\textwidth]{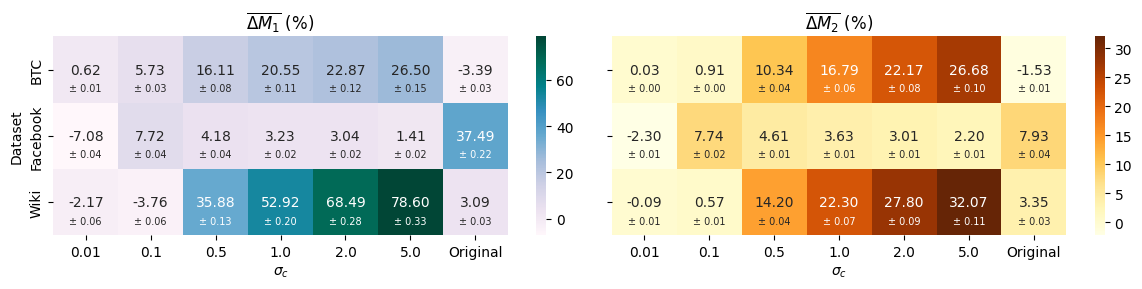}
    \label{fig:dmon_fcom_dice_real_networks}
    }

    \subfloat[DiffPool]{
        \includegraphics[width=0.499\textwidth]{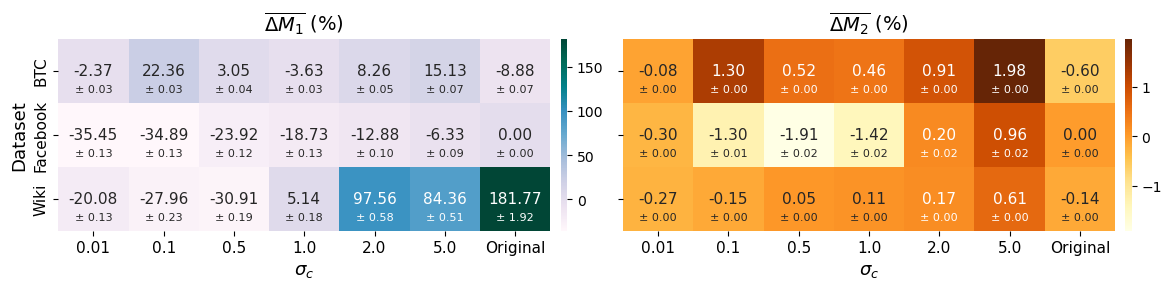}
        \label{fig:diffpool_fcom_dice_real_networks}
        }
    \subfloat[MinCut]{
        \includegraphics[width=0.499\textwidth]{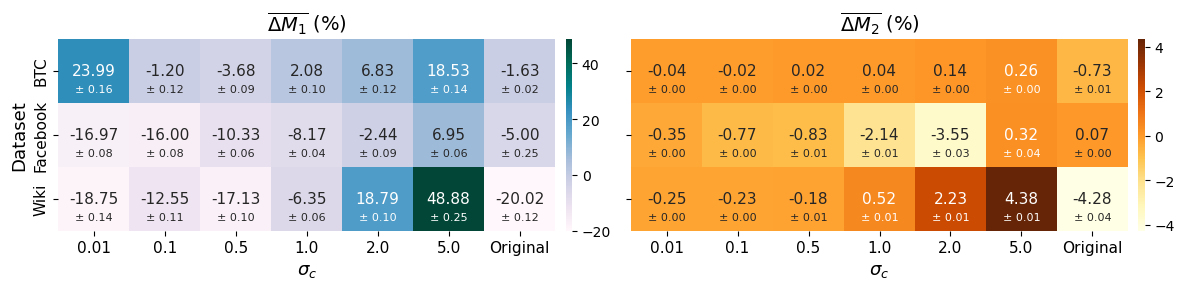}
        \label{fig:mincut_fcom_dice_real_networks}
        }
    \caption{Average relative improvement (\% with $\pm 1$ s.d.) of FCom-DICE over baseline DICE averaged over all $\beta_b$ and all realizations using (a) DMoN, (b) DiffPool and (c) MinCut for community detection. The relative performance difference is illustrated for real networks with node features produced by a multivariate Gaussian distribution with varying $\sigma_c$ (refer to Section~\ref{sec:features_LFR} for methodological details) and original features (right column).}
    \label{fig:delta_fcom-dice_real_networks}
\end{figure*}

Figure~\ref{fig:delta_fcom-dice_real_networks} reports the relative improvement of FCom-DICE over DICE in terms of $\overline{\Delta M_1}(\texttt{Dataset}, \sigma_c)$ and $\overline{\Delta M_2}(\texttt{Dataset}, \sigma_c)$, across different values of $\sigma_c$, where $\sigma_c$ controls the variance of node features generated from a multivariate Gaussian distribution (see Section~\ref{sec:features_LFR} for detailed methodology).

For DMoN, FCom-DICE consistently improves concealment over structure-only DICE as $\sigma_c$ increases. The effect is most pronounced on Wikipedia, where higher $\sigma_c$ features lead to large relative gains in both metrics. On Bitcoin Transactions, improvements are moderate, which could be reflecting the sparse and weakly modular nature of the network. Facebook exhibits the smallest overall improvement across all $\sigma_c$ values (most improvement at $\sigma_c = 0.1$). However, with original features, the results for the Facebook network are considerably higher for $M_1$. 
% This behavior could be explained by the higher density of edges in the network (see Table~\ref{tab:real_networks_stats} of Appendix~\ref{appx:real_datasets}). 

The results for DiffPool and MinCut are less stable, which is consistent with their weaker community recovery compared to DMoN (discussed in Section~\ref{sec:results_diffpool_mincut}).
For DiffPool, FCom-DICE still improves $M_1$ in several settings, especially for Wikipedia at high $\sigma_c$ and with original features, but the gains in $M_2$ are small and sometimes negative. 
% Facebook is negative
MinCut shows a similar pattern: improvements are mostly concentrated in $M_1$, while $M_2$ remains close to zero or negative for Facebook and Bitcoin Transactions. Note that for DiffPool and MinCut, the baseline concealment is often already high because the detected partitions are coarse or near-trivial. For example, on Facebook, $M_1$ is often close to $0$ and $M_2$ is close to $1$, meaning that the target community is already poorly recovered by the detector. Therefore, FCom-DICE has limited room to further improve concealment, and small changes in the detected partitions can appear as negative relative gains. Thus, the smaller and less consistent gains for DiffPool and MinCut mostly reflect the instability of these models rather that a systematic failure of FCom-DICE.

For low values of $\sigma_c$, corresponding to overlapping and noisy features, we observe little to no improvement, and in some cases small negative relative gain. 
This is aligned with findings on LFR, indicating that with noisy features, FCom-DICE offers limited additional leverage beyond structure-only attacks.

\subsection{Utility Preservation after FCom-DICE}
In this section, we evaluate whether FCom-DICE preserves the utility of the perturbed graph by examining changes in its global structural and feature characteristics. Specifically, we assess the preservation of the degree distribution, community structure, and node-feature distribution after perturbation.

\subsubsection{Degree Distribution}
The degree distribution is a fundamental structural characteristic of a graph. Many real-world networks exhibit heavy-tailed, approximately power-law degree distributions~\cite{clauset_power-law_2009}. Similarly, the LFR benchmark used in our experiments generates node degrees according to a power-law distribution.

To assess whether the perturbed graph $G'$ preserves the degree-distribution characteristics of the original graph $G$, we use the two-sample likelihood-ratio test for power-law distributions proposed by Bessi~\cite{bessi_two_2015} and adopted in Nettack~\cite{zugner_adversarial_2018-1}. 
The test compares a null model in which $G$ and $G'$ share the same power-law exponent with an alternative in which their exponents differ.
The exponent $\alpha$ of the power-law distribution $p(x)\propto x^{-\alpha}$ is estimated using the approximate maximum-likelihood estimator introduced by Clauset et al.~\cite {clauset_power-law_2009} for discrete data. Following the Nettack~\cite{zugner_adversarial_2018-1} implementation, we consider degrees ($d\geq d_{\min}$), with ($d_{\min}=2$).

Across all experimental realizations described in Table~\ref{tab:experiments_params} (see Section~\ref{sec:lfr}), the likelihood-ratio test fails to reject the null hypothesis that the original and perturbed degree sequences share a common power-law exponent ($p>0.05$). Additionally, Zugner et al.~\cite{zugner_adversarial_2018-1} define changes to the degree distribution as \textit{unnoticeable} when the likelihood-ratio statistic $\Lambda (G, G') < 0.004$, which approximately corresponds to a p-value of $0.95$. FCom-DICE satisfies this criterion for all cases with budgets ($\beta_b$) up to and including 15\% of the target community’s intra-community edges. 
Above this budget, unnoticeability threshold violations occur for large target communities in graphs with strong structural separation, particularly for $\mu = 0.01$ or $0.1$. The maximum test statistic $\Lambda = 0.07$ (p-value $0.79 \geq 0.05$), obtained at the perturbation budget $\beta_b = 100\%$  for a target community of size $100$ nodes with $\mu=0.01$. This suggests that the proposed approach, FCom-DICE, achieves degree-distribution unnoticeability with a perturbation budget $\beta_b \leq 15\%$, while remaining statistically drawn from the same power-law distribution as the original graph $G$ across all evaluated settings, thereby preserving utility.

\subsubsection{Community Structure}
\label{appx:fcom-dice_ecs}
We quantify global community structure similarity between the original and perturbed graphs using the Element-Centric Similarity (ECS)~\cite{gates_element-centric_2019}, which has been used before to compare the community structures before and after the network perturbation and test the robustness of the community structure~\cite{wei_robustness_2024, goel_community_2026}. 

\begin{figure*}[!hbtp]
    \centering
    \includegraphics[width=0.9\textwidth]{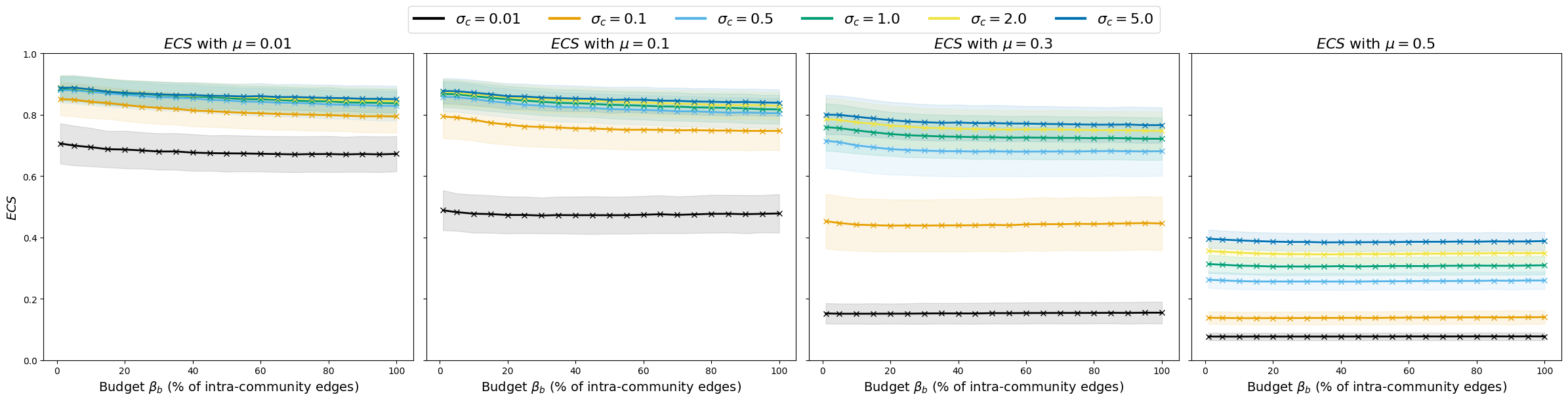}
    \caption{ECS of \textit{FCom-DICE} across $\sigma_c$, $\mu$, and perturbation budget $\beta_b$ ($p=0.5$). Shaded bands denote $\pm 1$ s.d. across runs.}
    \label{fig:ecs}
\end{figure*}

Figure~\ref{fig:ecs} shows that ECS remains largely stable across the full range of perturbation budgets $\beta_b$ for all tested $\mu$ and $\sigma_c$ values. Although the absolute ECS level shifts with different $\mu$ and $\sigma_c$, the curves are nearly flat with respect to $\beta_b$, indicating that increasing the budget does not create structural distortion in the graph. 
This suggests that even at high perturbation levels, the overall network community structure, as captured by ECS, remains largely preserved and that the subsequent effects on concealment metrics are not driven by a general breakdown of connectivity.

\subsubsection{Node Feature Distribution}
% As was discussed in Section~\ref{sec:features_LFR}, we featurize LFR for our experiments using the process based on multivariate Gaussian distributions, such that a ground truth community $C_i^{\text{LFR}} \equiv C^f_i$, a cluster in feature space. 
% We evaluate whether the FCom-DICE method of concealing a community introduces a statistically detectable deviation from the feature-generation process underlying the original feature matrix $\mathbf X$.
We evaluate whether FCom-DICE introduces a statistically detectable deviation from the process used to generate the original node feature matrix $\mathbf X$.
To quantify the discrepancy between the distributions of the original feature matrix $\mathbf{X}$ and the perturbed one $\mathbf{X'}$, we use maximum mean discrepancy (MMD)~\cite{gretton_kernel_2012}.
The computation involves Gaussian RBF kernel $k(\mathbf x,\mathbf y)= \exp \left(-\gamma \bigl\|\mathbf x-\mathbf y\bigr\|_2^2\right)$, where $\gamma$ is determined using the median heuristic applied to the pairwise distances in the original feature matrix $\mathbf{X}$~\cite{gretton_kernel_2012}. $\gamma$ is fixed for both the observed statistic $ T_{\mathrm{obs}} = \widehat{\operatorname{MMD}}^2 \left( \mathbf X,\mathbf X' \right)$ and the simulated null distribution (described below). 

The Gretton et al. two-sample MMD test uses random permutations to estimate the null distribution, assuming independent samples under the null hypothesis~\cite{gretton_kernel_2012}. We cannot assume this in our setting because $\mathbf X'$ is obtained by perturbing $\mathbf X$. 
Instead, we use a parametric Monte Carlo null distribution approximation since the synthetic feature-generation process is known. 
% , and whether $\mathbf X'$ is significantly different from $\mathbf X$ in the distribution process, we c. 
% using the same parameters used to produce the original synthetic node features (see Sections~\ref{sec:features_LFR} and Table~\ref{tab:experiments_params}). 
We generate $1,000$ independent feature matrices for each graph using the Gaussian feature-generation process described in Section\ref{sec:features_LFR}, preserving all parameters used in $\mathbf X$, such as community assignments, community centroids, and feature centroids separation.
We calculate $T_b = \widehat{\operatorname{MMD}}^2 \left( \mathbf X, X^{(b)} \right), b=1,\ldots,1000$ for each realization.
The $\{T_1,\ldots,T_{1000}\}$ represent an empirical approximation of the null distribution.
For each graph, we estimate the empirical 95th percentile ($q_{0.95}$) and reject the null hypothesis when $T_{\mathrm{obs}} > q_{0.95}$. 
Thus, the test determines whether the observed MMD exceeds the variability expected under the original feature-generation process.

\begin{figure}[!htbp]
    \centering
    \includegraphics[width=0.7\linewidth]{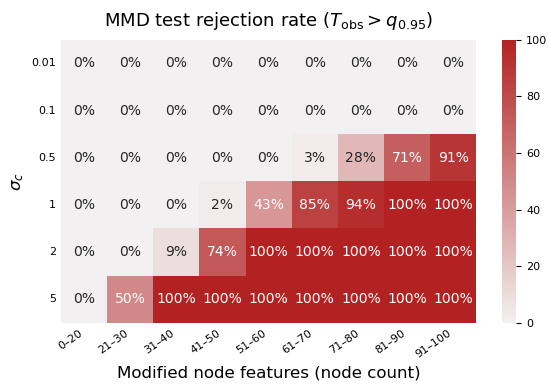}
    \caption{Percentage of experimental realizations for which the MMD test rejected the null distribution at the 5\% significance level, grouped by feature separation $\sigma_c$ and the realized number of modified node features.}
    \vspace{-1em}
    \label{fig:MMD_test}
\end{figure}

Figure~\ref{fig:MMD_test} shows the MMD test rejection rate across $\sigma_c$ and the realized number of modified node features. We report the realized modifications rather than the perturbation budget $\beta_b$, as the MMD statistic directly reflects changes to the feature matrix.
% As the number of modified nodes increases, perturbations in the feature matrix become increasingly distinguishable from the original. However, when feature separation is small (i.e., $\sigma_c \leq 0.1$), the perturbations remain statistically indistinguishable. However, as $\sigma_c$ increases, changes in the feature matrix introduced by the FCom-DICE concealment method become more detectable. For well-separated communities ($\sigma_c = 5$), modifying as few as $21-30$ nodes (out of $1{,}000$) is sufficient for the test to reject the null distribution in approximately half of the realizations. 
Detectability increases with both the number of modified nodes and feature separation. For $\sigma_c \leq 0.1$, perturbations remain statistically indistinguishable, whereas at $\sigma_c=5$, modifying only $21$--$30$ of $1{,}000$ nodes leads to rejection in approximately half of the realizations.

\section{Conclusion and Discussion}
\label{sec:conclusion}
% This paper addresses a group-level privacy risk arising from GNN-based community detection and studies how a targeted community can be concealed from it. 
This work studies targeted community concealment in attributed networks under GNN-based community detection. 
% Our work was inspired by literature documenting the importance of group privacy for personal safety and political freedom~\cite{bonneau_eight_2009}.
Our analysis shows that two factors majorly govern hidability: (i) the ratio of external to internal edges and (ii) the feature-space proximity of the target community to other communities. 
Guided by these findings, we proposed \emph{FCom-DICE}, a feature-aware extension of DICE that deletes intra-community edges, adds cross-community edges toward the most feature-similar community, and applies feature modifications to match the linked community. 
Across featurized LFR graphs with ground truth evaluated with DMoN, FCom-DICE consistently outperforms structure-only baseline DICE and reaches strong concealment at smaller perturbation budgets. FCom-DICE also performs better than the baseline on featurized real networks, particularly when features are distinct to each community. However, FCom-DICE and DICE perform similarly %on featurized real networks does not perform better than a baseline 
when features overlap significantly or are excessively noisy.
Additional experiments with DiffPool and MinCut show that FCom-DICE generally improves over DICE beyond DMoN, although the gains are smaller and less stable due to weaker overall community recovery with these models. Thus, the effectiveness of the proposed feature-guided perturbation approach is strongest when the underlying detector can strongly recover community structure using both the community structure and feature separability.

% \noindent\textbf{Implications.}
% From a privacy perspective, our findings suggest two complementary strategies for protecting high-value communities: limiting the formation of dense, internally isolated clusters (enhancing cross-community ties) and reducing feature distinctiveness relative to other communities.
% These results highlight how collective structure and attributes jointly contribute to group-level privacy leakage in graph learning pipelines.

% Conversely, defenders aiming to identify concealed adversarial clusters or communities within the network should focus on structural and feature signals, such as abrupt increases in external/internal edge ratios for a community or feature drifts towards other communities. Further research is necessary to investigate techniques that can enhance the robustness of GNN methods against this issue. 

These results highlight that the interaction between structural and feature separability governs community concealment from GNN-based methods. Within the evaluated setting, feature-guided perturbations provide additional improvements over structure-only perturbations, especially when node features reinforce community clustering. This pattern is reflected in the results, where FCom-DICE improves concealment significantly over DICE in highly separable features. 

The comparison across DMoN, DiffPool, and MinCut further shows that concealment results also depend on the robustness of the underlying community detector. When the community detection method recovers meaningful partitions close to the ground truth, FCom-DICE yields clear improvements; however, when the detector produces near-trivial partitions, the baseline concealment is already high, and relative improvement is not noticed.

As was highlighted previously, the goal of the paper is to study the mechanisms that govern concealment of a target community. To do that, we use synthetically generated networks with node features, specifically to control and experiment with different conditions in a network. This requires a benchmark with known ground-truth communities and independently controlled structural properties. LFR remains a major benchmark that provides a realistic power-law degree distribution, heterogeneous community sizes, and explicit control over structural noise (mixing parameter $\mu$). Thus, a great fit for this study.
We additionally generate features for nodes from Gaussian distributions to independently control feature separability while holding graph structure fixed. This enables the investigation of how node features influence community concealment under GNN-based community detection. More realistic features or real-world features introduce additional confounding factors that can make it difficult to control and observe the effect on community concealment.

\noindent\textbf{Limitations.}
We advise researchers to consider some constraints. 
First, a Gaussian distribution with community-defined clusters generates synthetic node features for experiment control. The experiments isolate the effect of feature separability by assuming features align with community structure, but they do not capture feature distribution diversity. Alternative featurization methods should be investigated, including cases where node features do not correlate with ground truth communities or follow domain-specific feature distributions. 
Second, since edge addition changes node features, the perturbation budget $b$ only accounts for edge perturbations while including feature perturbations. Thus, we assume it is a perturbation subprocess. Studies on feature perturbation as a GNN attack often manipulate features without affecting edge perturbations~\cite{zugner_adversarial_2018, xu_attacks_2024}. Future work should model feature perturbation costs and plausibility constraints separately.
Third, our method FCom-DICE assumes the defender trying to hide a high-value community has network knowledge, including community structure and features. Exploring scenarios involving partial or complete absence of such knowledge is a promising avenue.

% Finally, operational deployments must balance concealment with functionality and detectability constraints (e.g., degree distributions, community size, attribute semantics); incorporating such ``unnoticability'' constraints on the budget is a key next step.

% out method requires to know the ground truth of the communiy labels of the entire network.
% DMoN requires number of classes

\appendices
% \section{Appendix}
\label{sec:appendix}

\section{Details on DICE Performance}
\label{appx:dice_details}
\begin{figure}[!hbtp]
    \centering
    \includegraphics[width=\linewidth]{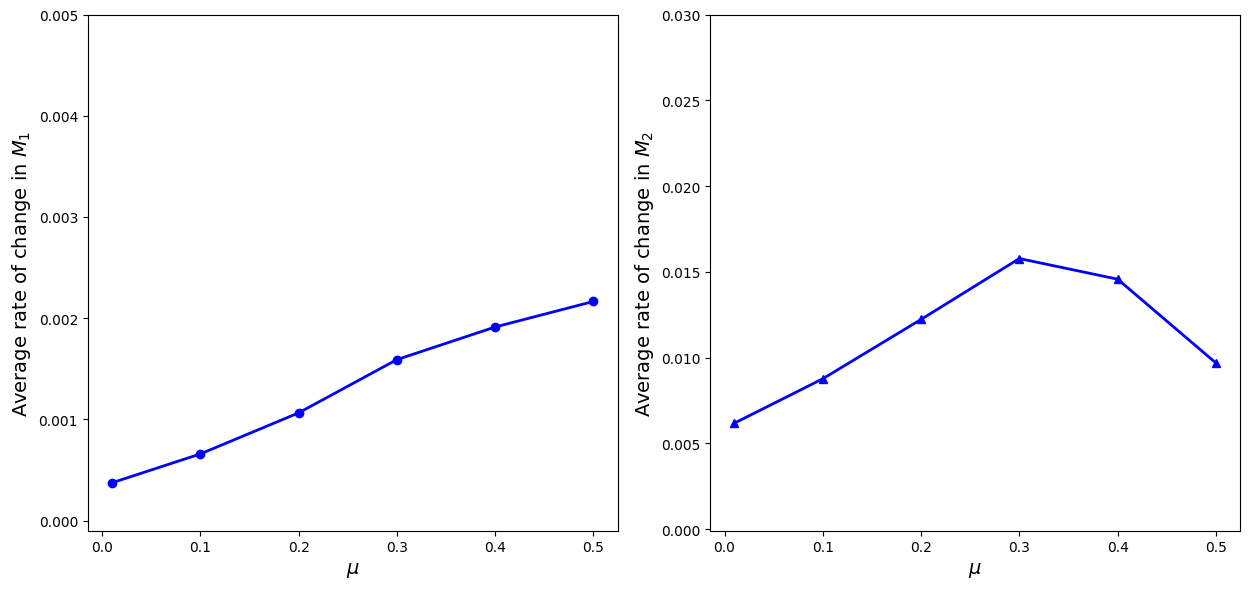}
    \vspace{-2em}
    \caption{Average rate of change of $M_1$ and $M_2$ vs. $\mu$, averaged over $\sigma_c$.}
    \label{fig:avg_change_mu}
\end{figure}

Figure \ref{fig:avg_change_mu} displays the average rate of change relative to budget as a function of $\mu$, averaged across all $\sigma_c$, highlighting the isolated impact of structural noise. The average rate of change in $M_1$ increases with $\mu$, while $M_2$ peaks at $\mu = 0.3$ before slightly decreasing. 

\section{Comparison of FCom-DICE with DICE}
\label{appx:fcom-dice_vs_dice}
% Our proposed method, FCom-DICE, yields a noticeable performance improvement over DICE (baseline). In Figure~\ref{fig:fcom-dice}, both $M_1$ (top row) and $M_2$ (bottom row) still increase with the perturbation budget $\beta_b$, and at the same time the overall trend of growing hidability with higher $\mu$ and lower $\sigma_c$ still holds. 
% However, with FCom-DICE, performance increases faster, particularly for larger $\mu$ (e.g., $\mu =0.3$ or $0.5$), reaching a plateau value only at perturbation budget $\beta_b \ 20\%$  , rather than gradually increasing across the entire budget range (as DICE in Figure~\ref {fig:dice}).
% For instance, at $\mu=0.5$ a low $\sigma_c$, $M_2$ rises to $\approx 0.8$ after less than 20\% of the budget, suggesting that the cap of the hiding capacity is achieved with a smaller perturbation compared to DICE.

\begin{figure*}[!htbp]
    \centering
    \includegraphics[width=0.9\textwidth]{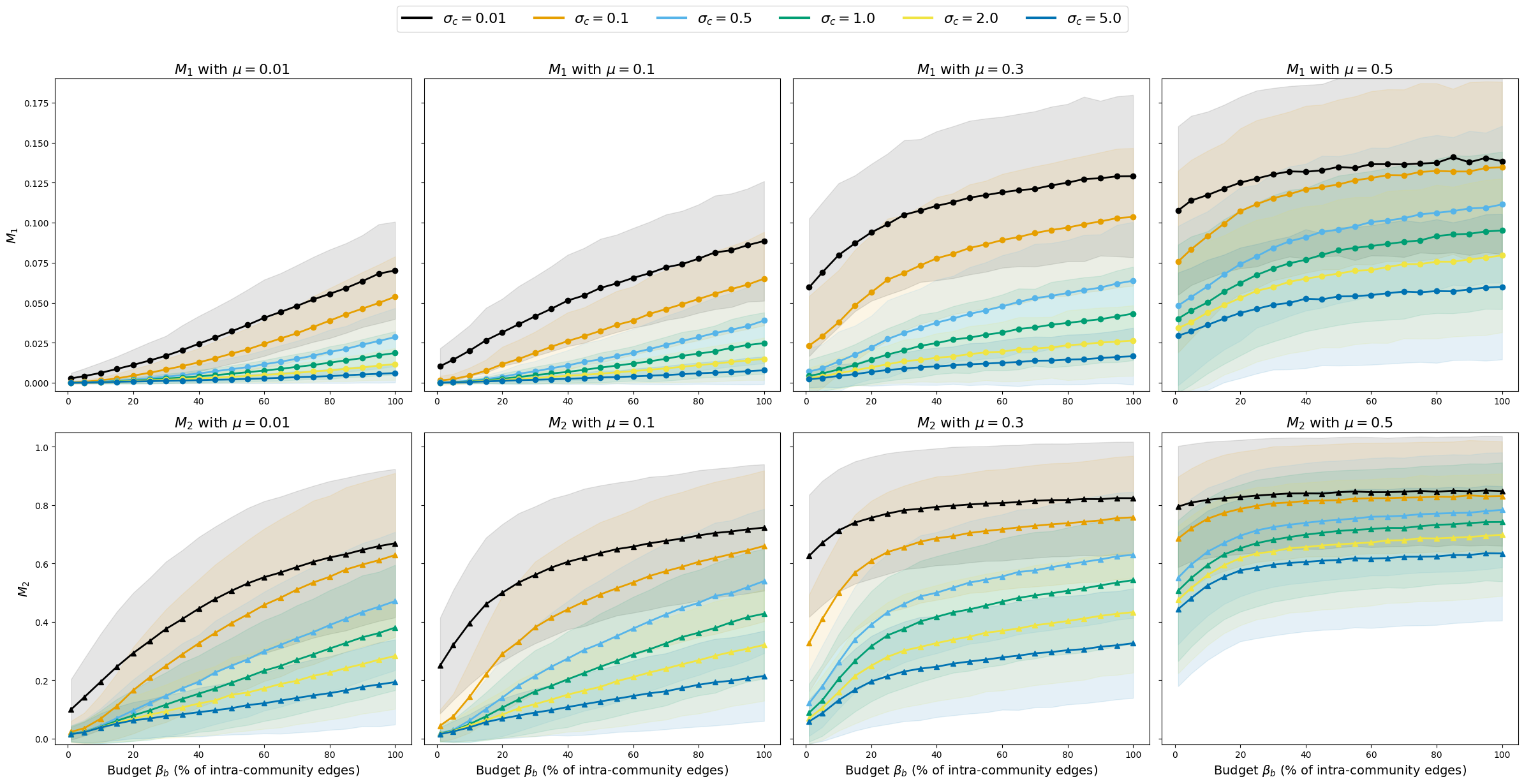}
    \caption{\textit{FCom-DICE} performance across $\sigma_c$, $\mu$, and perturbation budget $\beta_b$ ($p=0.5$). Shaded bands denote $\pm 1$ s.d. across runs. For comparison with DICE see Figure~\ref{fig:dice}}
    \label{fig:fcom-dice}
\end{figure*}

The proposed method, \textit{FCom-DICE}, outperforms DICE (baseline, see Figure~\ref{fig:dice}). Figure\ref{fig:fcom-dice} shows that both $M_1$ (top row) and $M_2$ (bottom row) increase with perturbation budget $\beta_b$, while concealment increases with higher $\mu$ and lower $\sigma_c$. 
Compared to DICE, \textit{FCom-DICE} exhibits faster performance increases, especially for larger $\mu$ (e.g., $\mu =0.3$ or $0.5$), reaching a plateau at a lower perturbation budget (around 20-30\%) (Fig.~\ref{fig:dice}). 
At $\mu=0.5$ and low $\sigma_c$, $M_2$ reaches $\approx 0.8$ after less than 20\% of the budget, indicating a smaller perturbation than DICE for achieving the concealment capacity cap.

% \begin{figure*}[!htbp]
%   \centering
%   \subfloat[\textit{FCom-DICE}.\label{fig:fcom-dice}]{ \includegraphics[width=\textwidth]{./figures/fcomdice_per_mu.png} 
%   }
%   % \begin{subfigure}{\textwidth}
%   %   \centering
%   %   \includegraphics[width=\textwidth]{./figures/fcomdice_per_mu.png}
%   %   \caption{\textit{FCom-DICE}.}
%   %   \label{fig:fcom-dice}
%   % \end{subfigure}

%   \vspace{0.5em}
%     \subfloat[DICE (baseline).\label{fig:dice_repeated}]{ \includegraphics[width=\textwidth]{./figures/dice_per_mu.png} 
%     }
%   % \begin{subfigure}{\textwidth}
%   %   \centering
%   %   \includegraphics[width=\textwidth]{.//figures/dice_per_mu.png}
%   %   \caption{DICE (baseline).}
%   %   \label{fig:dice_repeated}
%   % \end{subfigure}

%   \caption{Comparison of \textit{FCom-DICE} and DICE performance across $\sigma_c$, $\mu$, and perturbation budget $\beta_b$ ($p=0.5$). Shaded bands denote $\pm 1$ s.d. across runs.}
%   \label{fig:fcom_vs_dice}
% \end{figure*}

% \section{Structural Similarity after FCom-DICE}

\section{Additional GNN Methods}
\label{sec:methods_other_gnn}

\subsection{DiffPool}
\label{sec:method_diffpool}
DiffPool learns a hierarchical graph representation by jointly learning node embeddings and a differentiable soft cluster assignment matrix~\cite{ying_hierarchical_2018}.
Given the node embedding matrix $\mathbf{H}^{(l)}$ and adjacency matrix $\mathbf{A}^{(l)}$ at layer $l$, DiffPool learns two GNN mappings: one for node embeddings and one for cluster assignments as follows:
\begin{align*}
    \mathbf{Z}^{(l)} 
    &= 
    \mathrm{GNN}^{(l)}_{\mathrm{embed}}
    \left(\mathbf{A}^{(l)}, \mathbf{H}^{(l)}\right), \\
    \mathbf{C}^{(l)} 
    &= 
    \mathrm{softmax}
    \left(
    \mathrm{GNN}^{(l)}_{\mathrm{pool}}
    \left(\mathbf{A}^{(l)}, \mathbf{H}^{(l)}\right)
    \right),
\end{align*}
% \vspace{-0.2em}
where $\mathbf{Z}^{(l)}$ denotes the learned node embeddings and $\mathbf{C}^{(l)}$ is the soft assignment matrix at layer $l$.
Then the pooled node embeddings and adjacency matrix are used as an input to layer $l+1$: 
$\mathbf{H}^{(l+1)} = {\mathbf{C}^{(l)}}^\top \mathbf{Z}^{(l)}$ and $\mathbf{A}^{(l+1)} = {\mathbf{C}^{(l)}}^\top \mathbf{A}^{(l)} \mathbf{C}^{(l)}$.
% \begin{align*}
%     \mathbf{H}^{(l+1)}
%     &=
%     {\mathbf{C}^{(l)}}^\top \mathbf{Z}^{(l)}, \\
%     \mathbf{A}^{(l+1)}
%     &=
%     {\mathbf{C}^{(l)}}^\top
%     \mathbf{A}^{(l)}
%     \mathbf{C}^{(l)}.
% \end{align*}

The unsupervised DiffPool objective is computed based on link-prediction loss and an entropy regularization: 
\begin{equation*}
\mathcal{L} = 
\underbrace{
    \left\|
    \mathbf{A}^{(l)} - \mathbf{C}^{(l)}{\mathbf{C}^{(l)}}^{\top}
    \right\|_F
}_{\mathcal{L}_{\mathrm{LP}}^{(l)}}
    +
\underbrace{
    \frac{1}{n^l}
    \sum_{i=1}^{n^l}
    H(\mathbf{C}^{(l)}_{i:})
}_{\mathcal{L}_{\mathrm{E}}^{(l)}},
\end{equation*}
where $H(\mathbf{C}^{(l)}_{i:})$ is the entropy of the assignment distribution for node $v_i$ at layer $l$. The final community assignment is obtained as: $\hat{y}_i = \arg\max_{c \in \{1,\dots,k\}} \mathbf{C}_{ic}$.
% \begin{equation*}
%     \hat{y}_i = \arg\max_{c \in \{1,\dots,k\}} \mathbf{C}_{ic}.
% \end{equation*}

\subsection{MinCut}
\label{sec:method_mincut}

MinCut is a spectral clustering-based pooling method that learns cluster assignments by optimizing a differentiable relaxation of the normalized min-cut function~\cite{bianchi_spectral_2020}.
% Given the node embedding matrix $\mathbf{H}^{(l)}$ and adjacency matrix $\mathbf{A}^{(l)}$ at layer $l$, MinCut learns cluster assignment matrix as follows:
% \begin{equation*}
%     \mathbf{C}^{(l)} =
%     \mathrm{softmax}
%     \left(
%     \mathrm{GNN}^{(l)}_{\mathrm{pool}}
%     \left(\mathbf{A}^{(l)}, \mathbf{H}^{(l)}\right)
%     \right),
% \end{equation*}

The main MinCut objective is composed of a cut loss and an orthogonality regularization:
\begin{equation*}
    \mathcal{L}
    = \underbrace{
    -
    \frac{
    \text{Tr}\left(\mathbf{C}^{\top}\mathbf{A}\mathbf{C}\right)
    }{
    \text{Tr}\left(\mathbf{C}^{\top}\mathbf{D}\mathbf{C}\right)
    }
    }_{\mathcal{L}_\mathrm{cut}}
    +
    \underbrace{
    \left\|
    \frac{\mathbf{C}^{\top}\mathbf{C}}
    {\left\|\mathbf{C}^{\top}\mathbf{C}\right\|_F}
    -
    \frac{\mathbf{I}_k}{\sqrt{k}}
    \right\|_F
    }_{\mathcal{L}_\mathrm{orth}},
\end{equation*}
where $\mathbf{D}$ is the diagonal degree matrix of $\mathbf{A}$,  $\mathbf{I}_k$ is the $k \times k$ identity matrix, $\mathbf{C} \in \mathbb{R}^{n \times k}$ is the cluster assignment matrix $\mathbf{C} = \mathrm{softmax}(\mathrm{GNN}(\mathbf{A}, \mathbf{H}))$, and $k$ is the number of clusters. 

The cut loss $\mathcal{L}_\mathrm{cut}$ encourages dense intra-cluster connectivity, while the orthogonality loss $\mathcal{L}_\mathrm{orth}$ discourages trivial cluster assignments.

\ifCLASSOPTIONcaptionsoff
  \newpage
\fi



\bibliographystyle{IEEEtran}
% \bibliography{dalya_references, references}
\bibliography{clean_ref, references}

%-------------------------------------------------------------------------------
% % use section* for acknowledgment
\ifCLASSOPTIONcompsoc
  \section*{Acknowledgments}
\else
  \section*{Acknowledgment}
\fi

% note for arxiv submission
This work has been submitted to the IEEE for possible publication. Copyright may be transferred without notice, after which this version may no longer be accessible.

\noindent This paper has been coauthored by UT-Battelle, LLC under Contract No.\ DE-AC05-00OR22725 with the U.S.\ Department of Energy. The publisher, by accepting the article for publication, acknowledges that the U.S.\ government retains a nonexclusive, paid up, irrevocable, world-wide license to publish or reproduce the published form of the manuscript, or allow others to do so, for U.S.\ government purposes. The DOE will provide public access to these results in accordance with the DOE Public Access Plan (\url{http://energy.gov/downloads/doe-public-access-plan}). This material is based upon work supported by the U.S. Department of Energy, Office of Science, Office of Advanced Scientific Computing Research under Contract No. DE-AC05-00OR22725. The funders had no role in study design, data collection and analysis, decision to publish, or preparation of this manuscript.

\noindent The work is also supported by the U.S. Department of Homeland Security award \#17STQAC00001-07-00 and NSF Grant No. TI 2449402. 
Any opinions, findings, and conclusions or recommendations expressed in this material are those of the author(s) and do not necessarily reflect the views of the National Science Foundation nor the Department of Homeland Security.

\noindent Experiments were conducted using HPC resources provided by the Indiana University Pervasive Technology Institute (supported in part by the Lilly Endowment, Inc.).

\end{document}